\documentclass[authoryear,preprint,review,12pt]{elsarticle}

\usepackage{amssymb}
\usepackage{amsmath}
\usepackage{amsthm}
\usepackage{amsfonts}
\usepackage{algorithmic}
\usepackage{algorithm}
\usepackage{array}
\usepackage{textcomp}
\usepackage{stfloats}
\usepackage{url}
\usepackage{verbatim}
\usepackage{graphicx}
\usepackage{cite}
\usepackage{siunitx}
\usepackage{booktabs}
\usepackage{threeparttable}
\usepackage{subcaption}
\usepackage{placeins}
\usepackage{eurosym}

\usepackage{xcolor}
\newcommand{\rev}[2]{{\color{#1} #2}}
\newcommand{\revd}[1]{\rev{black}{#1}}

\journal{Engineering Applications of Artificial Intelligence}

\begin{document}

\begin{frontmatter}

\title{Sim-to-Real Transfer via a Style-Identified Cycle Consistent Generative Adversarial Network: Zero-Shot Deployment on Robotic Manipulators through Visual Domain Adaptation}

\author[1]{Luc\'{i}a G\"{u}itta-L\'{o}pez \corref{cor1}}
\ead{lucia.guitta@iit.comillas.edu}

\author[2,3]{Lionel G\"{u}itta-L\'{o}pez}
\ead{lglopez@comillas.edu}

\author[1]{Jaime Boal}
\ead{jaime.boal@iit.comillas.edu}

\author[1]{\'{A}lvaro J. L\'{o}pez-L\'{o}pez}
\ead{alvaro.lopez@iit.comillas.edu}

\cortext[cor1]{Corresponding author}

\address[1]{Institute for Research in Technology~(IIT), ICAI School of Engineering, Comillas Pontifical University,
			Rey Francisco, 4,
			28008,
			Madrid,
			Spain}
\address[2]{ICAI School of Engineering, Comillas Pontifical University,
			Alberto Aguilera, 25,
			28015,
			Madrid,
			Spain}
\address[3]{Gestamp,
			Omb\'{u}, 3,
			28045,
			Madrid,
			Spain}

\begin{abstract}
The sample efficiency challenge in Deep Reinforcement Learning (DRL) compromises its industrial adoption due to the high cost and time demands of real-world training. Virtual environments offer a cost-effective alternative for training DRL agents, but the transfer of learned policies to real setups is hindered by the sim-to-real gap. Achieving zero-shot transfer, where agents perform directly in real environments without additional tuning, is particularly desirable for its efficiency and practical value. This work proposes a novel domain adaptation approach relying on a Style-Identified Cycle Consistent Generative Adversarial Network (StyleID-CycleGAN or SICGAN), an original Cycle Consistent Generative Adversarial Network (CycleGAN) based model. SICGAN translates raw virtual observations into real-synthetic images, creating a hybrid domain for training DRL agents that combines virtual dynamics with real-like visual inputs. Following virtual training, the agent can be directly deployed, bypassing the need for real-world training. The pipeline is validated with two distinct industrial robots in the approaching phase of a pick-and-place operation. In virtual environments agents achieve success rates of 90 to 100\%, and real-world deployment confirms robust zero-shot transfer (i.e., without additional training in the physical environment) with accuracies above 95\% for most workspace regions. We use augmented reality targets to improve the evaluation process efficiency, and experimentally demonstrate that the agent successfully generalizes to real objects of varying colors and shapes, including LEGO\textsuperscript{\textregistered}~cubes and a mug. These results establish the proposed pipeline as an efficient, scalable solution to the sim-to-real problem. 

This article was accepted and published in Engineering Applications of Artificial Intelligence. Cite as: G\"{u}itta-L\'{o}pez, L., G\"{u}itta-L\'{o}pez, L., Boal, J. \&  L\'{o}pez-L\'{o}pez, A.J. Sim-to-real transfer via a Style-Identified Cycle Consistent Generative Adversarial Network: Zero-shot deployment on robotic manipulators through visual domain adaptation. Eng. Appl. Artif. Intell. 159, PA (Nov 2025). \url{https://doi.org/10.1016/j.engappai.2025.111510}
\end{abstract}


\end{frontmatter}



\section{Introduction}
\label{Intro}
Industrial robotic manipulators primarily perform repetitive tasks in controlled environments, limiting their applicability. Enhancing their ability to operate under dynamic conditions requiring sequential decision-making could greatly expand their use across various industries~\citep{Kobbi2025, Cho2025, Gai2024}. To address this challenge, Deep Reinforcement Learning (DRL)~\citep{Sutton2018, Francois-Lavet2018, Matsuo2022} has emerged as a prominent machine learning method, capable of handling complex problems with high-dimensional state spaces, such as images, where an agent interacts with a dynamic environment and performs actions sequentially~\citep{Bengio2013}. Leveraging images as the primary data source not only enables an asset-independent methodology but also allows robotic manipulators to adapt to changing environments and integrate advanced methods within a unified framework.

Despite its advantages, a major drawback of DRL algorithms is the substantial amount of experience (i.e., interactions between the agent and the environment) required for effective learning~\citep{Francois-Lavet2018}. One efficient solution to this \textit{sample efficiency} problem is to train the agent in a virtual environment. The virtual setup accelerates learning while reducing financial and time costs associated with real assets. Once the agent learns in the virtual environment (source domain), the challenge shifts to efficiently transferring the learned knowledge to the real environment (target domain). Bridging the reality gap depends on the discrepancies between the two domains~\citep{Zhu2023}. The most efficient approach to this \textit{sim-to-real} problem is \textit{zero-shot transfer}, where the agent trained in the source domain is directly deployed in the target domain without requiring additional fine-tuning using target-domain experience. Effective zero-shot transfer typically relies on maximizing the invariance of the learned policy in the source domain to visual and dynamic discrepancies that may arise in the target domain.

\begin{figure}[h!]
    \centering
    \includegraphics[width=\linewidth]{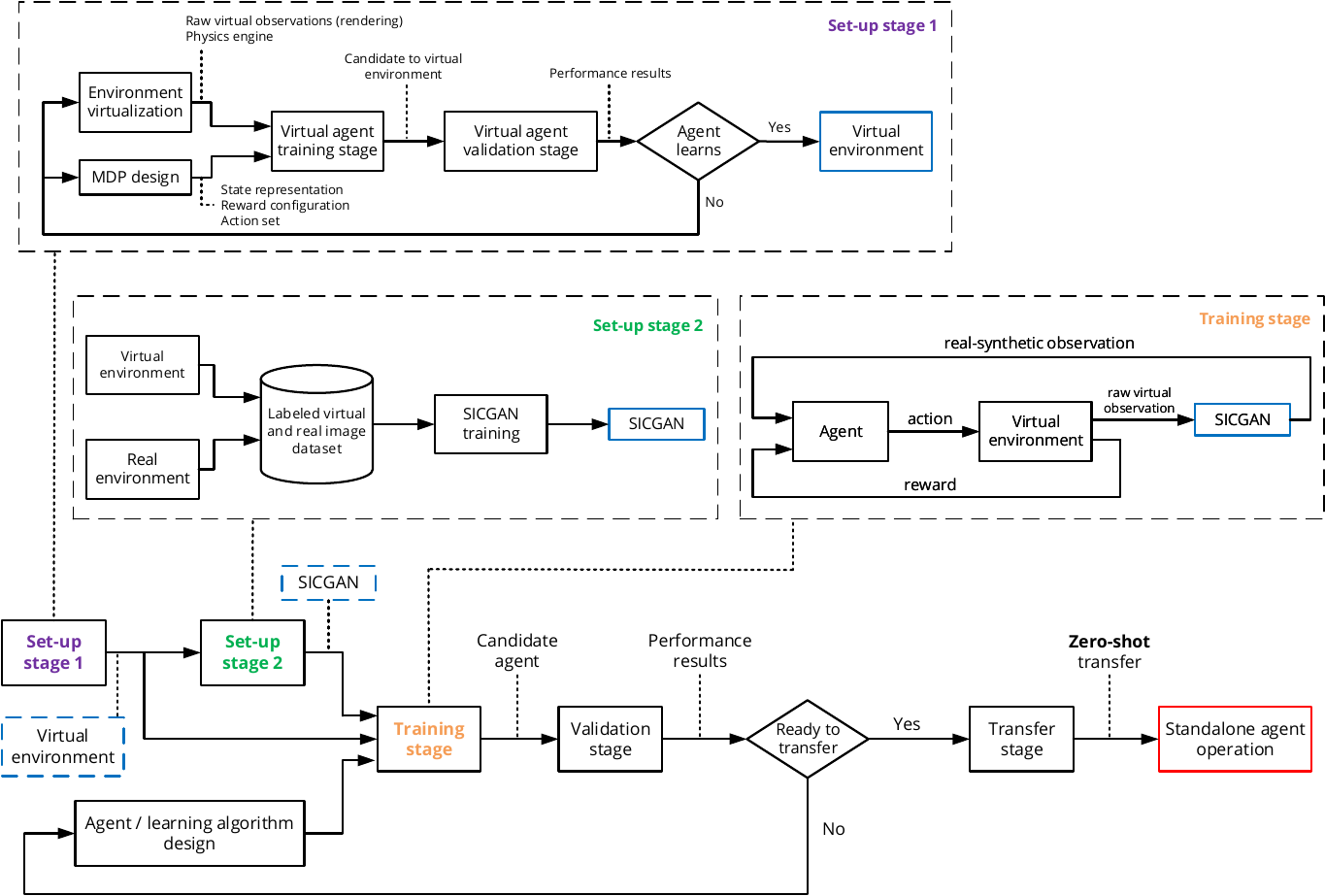}
    \caption{General proposed pipeline for sim-to-real zero-shot transfer in industrial applications. The pipeline begins with stage 1, where the virtual environment is created, and the MDP is defined. Stage 2 outlines the steps required to train the SICGAN. Using this SICGAN, the DRL agent is trained using real-synthetic observations generated from raw virtual data. Once the agent is trained and evaluated in the virtual environment, it is deployed directly to the real environment via zero-shot transfer.}
    \label{fig:generalPipeline}
\end{figure}

In this paper, we propose a Domain Adaptation (DA)~\citep{Wang2018} solution that uses a novel implementation of the Cycle Consistent Generative Adversarial Network (CycleGAN)~\citep{Zhu2017}, termed Style-Identified Cycle Consistent Generative Adversarial Network (StyleID-CycleGAN or SICGAN), to translate images from the source domain to the target domain. To assess the benefits of our DA architecture compared to other methods, we provide a comparison between our SICGAN and the original CycleGAN, as well as a state-of-the-art method based on visual transformers. We then train a DRL agent in the virtual environment using these translated observations (i.e., real-synthetic images) to enable zero-shot transfer in the real environment. Figure~\ref{fig:generalPipeline} presents the general overview of the steps followed from the environment virtualization to the zero-shot deployment in the real setup. Specifically, we demonstrate this approach with a 6-DoF industrial robotic arm whose goal is to approach a target randomly positioned within its workspace. This task represents the initial phase of a pick-and-place operation for any robotic manipulator. Our problem's complexity arises from the random relocation of the target between episodes and the reliance on the RGB environment image as the sole state observation, without proprioceptive information. Consequently, the agent must learn the state representation from a 64$\times$64$\times$3 image observation. Using images enhances the agent’s generalization capabilities, improves the versatility of the proposed pipeline, and facilitates the integration of additional image-based tasks.

The main contributions of this paper are the following:
\begin{itemize}
\item{The design and successful validation of a comprehensive pipeline that addresses the sim-to-real gap in an industrial robotic setting via zero-shot transfer, integrating an original DA model for image-to-image translation and a DRL agent trained in a virtual environment with translated observations. Validation is conducted using a different robotic manipulator from the one initially used to define the pipeline.}
\item{An original CycleGAN-based implementation, named StyleID-CycleGAN (SICGAN), which incorporates improvements from similar architectures and reduces the prevalence of artifacts commonly observed in the standard CycleGAN model.}
\end{itemize}

The remainder of the paper is organized as follows: Section~\ref{RW} situates our work within the current state-of-the-art research on this topic. Section~\ref{PD_ES} defines the problem solved and describes the experimental setups (i.e., the virtual and real environments used in our experiments). Section~\ref{Method} provides a detailed description of the proposed pipeline, implementation details, and validation procedure. Section~\ref{Results_Discussion} presents the results and discusses the outcomes, starting with the comparison between SICGAN with the original CycleGAN and a state-of-the-art visual DA method, and followed by an analysis of performance in the virtual setup and during zero-shot transfer. Finally, Section~\ref{Conclusion} summarizes the key findings of our research and briefly outlines future work.

\section{Related work}
\label{RW}
The use of DRL agents trained in virtual environments addresses the sample efficiency problem but requires techniques to transfer this virtual knowledge to real-world applications, as the disparity between domains widens the \emph{sim-to-real} gap. Several transfer learning methods exist to bridge this gap~\citep{Zhu2023}. \emph{Reward shaping} modifies the reward function in the target domain using external knowledge to provide auxiliary rewards that guide learning. \emph{Policy transfer} involves transferring policies through policy distillation, where knowledge from a teacher agent informs a student, or through policy reuse, applying source policies directly to the target task. \emph{Inter-task mapping} aligns state or action spaces across domains, enabling knowledge transfer for related but distinct tasks. Finally, \emph{representation transfer} leverages shared feature representations across domains, often using neural networks to learn invariant features.

Transfer learning can be categorized by the degree of refinement needed in the real environment. In \emph{zero-shot} transfer~\citep{Kirk2023}, the agent, trained solely in the virtual domain, is deployed directly in the real setting without additional target-specific experience. This approach assumes a high similarity between virtual and real observations and dynamics, achieved with photorealistic simulations or pre-processing techniques. \emph{Few-shot} learning refines the agent's performance with minimal real-world experience, often by partially retraining selected model layers. Finally, \emph{fine-tuning} involves extensive adjustment based on real experience, modifying model parameters to fit the target domain; while effective, this method requires substantial tuning efforts and computational resources.

Focusing on achieving zero-shot transfer, as it is the most efficient approach and is more likely to meet the common hardware constraints found in industrial settings, several techniques can be implemented. Among the various approaches --Domain Randomization (DR)~\citep{Tobin2017}, Domain Adaptation~\citep{Wang2018}, Knowledge Distillation~\citep{Rusu2016}, Progressive Neural Networks~\citep{Rusu2016_b}, Semantic Knowledge approaches, Meta RL~\citep{Wang2016}, or Imitation Learning~\citep{Hussein2017}-- DA appears to be the most suitable for the setup and problem addressed here. Unlike DR, which aims to train the agent in a virtual setting with a wide distribution of scenarios with randomized features so that the real observation is perceived as part of a known distribution, DA seeks to merge features from both the virtual and real domains to enable effective agent training within this hybrid domain. \revd{As a complementary technique to DA, Meta RL addresses generalization across tasks and domains, following a strategy based on rapid adaptation through task inference. As such, it can be seen as a complementary direction to DA. However, its increased algorithmic and computational complexity makes it more suitable for highly dynamic or multitask scenarios, rather than the constrained and efficiency-driven industrial contexts considered in this work.}

Some DA methods address the fusion of virtual and real domains by establishing a shared latent space that facilitates experience transfer between them. \citep{Shoeleh2020} applies DA in continuous RL by learning high-level skills in the source domain, which are then merged with target data to assist the agent in mapping state-action pairs across domains, thereby accelerating training in new tasks or configurations. Similarly, \citep{Higgins2017} introduces DARLA, a multi-stage RL agent that first learns a disentangled representation of the visual environment, then directly transfers policies from the source to the target domain. \citep{Xing2021} proposes a two-stage framework that develops a Latent Unified State Representation (LURS), consistent across domains, and integrates it with a source-domain RL policy. Along these lines, \citep{Li2022} introduces a two-stage model in which DA aligns domain-invariant state representations in a shared latent space, aiding the training of DRL agents. \citep{Chen2021} presents a cross-modal DA method (CODAS) that learns to map low-dimensional states from the source domain to high-resolution images in the target domain. \citep{Jeong2020} proposes a self-supervised sim-to-real approach using the temporal structure of DRL to adapt a simulation-trained latent representation to the real world.

Other DA approaches employ image-to-image translation between domains~\citep{Murez2018, Pang2022}. Generative Adversarial Networks (GAN)~\citep{Goodfellow2014} are frequently used for this visual adaptation. GANs utilize adversarial training between a generator and a discriminator to produce synthetic images indistinguishable from real samples. The generator ($G$) produces samples from a noise vector, which the discriminator ($D$) then classifies as real or synthetic, in a minimax game where $D$ maximizes and $G$ minimizes the objective.

Modifications to GANs address limitations in certain applications. For instance, CycleGAN~\citep{Zhu2017} was developed for unpaired data in image translation, employing two GANs, one per domain $\mathcal{X}$ and $\mathcal{Y}$, with an added cycle-consistency loss that ensures an image translated to the target domain and back remains unchanged as shown in~\eqref{eq:SOTA_TL_DA_CycleGANloss}.
\begin{equation}
\resizebox{.8\hsize}{!}{$
\begin{aligned}
\mathcal{L}_{\mathrm{cyc}} &= \mathcal{L}_{\mathrm{cyc}}^{G \rightarrow F} + \mathcal{L}_{\mathrm{cyc}}^{F \rightarrow G} \\
&= \mathbb{E}_{x \sim p_{\mathcal{X}}} \left[ \| F(G(x)) - x \|_1 \right] + \mathbb{E}_{y \sim p_{\mathcal{Y}}} \left[ \| G(F(y)) - y \|_1 \right]
\end{aligned}$}
\label{eq:SOTA_TL_DA_CycleGANloss}
\end{equation}
where $G: \mathcal{X} \rightarrow \mathcal{Y}$ and $F: \mathcal{Y} \rightarrow \mathcal{X}$ are the generators that map between the two domains $\mathcal{X}$ and $\mathcal{Y}$, and $x$ and $y$ are samples drawn from distributions $p_{\mathcal{X}}$ and $p_{\mathcal{Y}}$ respectively.

In order to address the limitations of the original CycleGAN, several architectures have emerged that build upon its foundational principles to varying degrees. These approaches aim to improve aspects such as visual fidelity, training stability, and semantic consistency in unpaired image-to-image translation tasks. Notable examples include UNet Vision Transformer cycle-consistent GAN (UVCGAN)~\citep{Torbunov2023} and its successor UVCGANv2~\citep{Torbunov2023_b}, which introduce uncertainty-guided consistency constraints and leverage stochastic augmentations to improve robustness and generalization. UVCGANv2 features a hybrid generator architecture that combines a U-Net backbone with an extended Vision Transformer (eViT) module at the bottleneck. The transformer receives the latent representation as a sequence of tokens, including a learnable style token that captures image-specific style information. This token is linearly transformed into style vectors that modulate the convolutional weights in the decoder via style-modulated convolutions, followed by demodulation to ensure activation stability. The discriminator includes a batch-statistics-aware head that uses cached features from previous batches to simulate large-batch diversity, enhancing training stability. Additionally, UVCGANv2 integrates training techniques such as spectral normalization, exponential moving average of generator weights, and a zero-centered gradient penalty to improve convergence and robustness. 

In contrast, our proposed SICGAN adopts a ResNet-based generator architecture and focuses on stabilizing training through demodulated convolutions and preserving semantic content using an identity loss. Further details on our implementation are described in Section~\ref{Method_SICGAN}. Unlike UVCGANv2, which relies on transformer-based style modulation, SICGAN emphasizes simplicity and efficiency, aiming for robust image translation in robotics contexts without significantly increasing architectural complexity or computational overhead. \revd{Comparing our SICGAN with the UVCGANv2 is particulary interesting because it highlights the trade-offs between architectural sophistication and practical deployment constraints. It also allows us to assess whether the benefits of transformer-based style control are critical for real-world performance in domains with strict efficiency requirements.}

In robotic grasping, \citep{James2019} proposes Randomized-to-Canonical Adaptation Networks (RCANs) to bridge the sim-to-real gap by translating randomized simulated images into canonical versions for zero-shot transfer with \qty{70}{\percent} accuracy, reaching \qty{91}{\percent} with few-shot fine-tuning. \citep{Ho2021} introduces RetinaGAN, enhancing CycleGAN with object structure consistency using EfficientDet~\citep{Tan2020}. For sim-to-real grasping, \citep{Bousmalis2018} combines DR with GraspGAN, achieving \qty{77}{\percent} accuracy. \citep{Rao2020} advances this work with RL-CycleGAN, using scene-consistency loss to retain task-relevant features, achieving \qty{95}{\percent} in single-bin grasping. For robotic assembly, \citep{Yuan2022, Shi2023} use CycleGAN for vision-based assembly, achieving \qty{86}{\percent} accuracy with latent-space representations, while \citep{Chen2022} combines sim-to-real and real-to-sim with CycleGAN and DR, obtaining \qty{92.3}{\percent}success but with camera pose limitations. In robot navigation, \citep{Zhang2019} reverses sim-to-real by translating real-world images back to the synthetic domain for continuity in the target domain. \citep{Jiang2021} introduces SimGAN to match simulated trajectories with real ones, and \citep{Jang2024} combines a modified CycleGAN with a shared latent space for successful real-world navigation.

Our approach primarily differs from previous works in that it fully considers hardware and software constraints, as well as economic and time limitations, inherent to industrial setups throughout the entire pipeline. Consequently, the resources used, along with training and evaluation processes, adhere strictly to these constraints. For example, our approach relies solely on a single externally mounted camera, eliminating the need for multiple cameras in the setup. Additionally, we have ensured that the designed solution is easily adaptable to other contexts, demonstrated by validating results using a different robot to perform the same task, which is not done in most of the previous works. With both robotic manipulators, the zero-shot transfer results generally exhibited outstanding results for most of the target objects evaluated.

\section{Problem definition and experimental setup}
\label{PD_ES}
\subsection{Problem definition}
\label{ES_ProbDef}
The task addressed is object approaching, representing the initial phase of common industrial operations like pick-and-place. Specifically, we focus on guiding a robotic arm close to an object randomly positioned within its workspace. This task is modeled as an episodic MDP where the initial positions of both the robot and the target are randomized at the episode beginning. The task is considered accomplished if the distance between the gripper and the object's center, a cube with \SI{6}{cm} sides, is less than \SI{5}{cm} during training and \SI{10}{cm} in evaluation. Training under stricter conditions (i.e., closer distance threshold) encourages better performance, a strategy commonly referred to as \textit{overlearning}. Each episode has a maximum length of 50 steps, ending early if the target is reached within the specified distance.

Although this task might appear straightforward for traditional control algorithms, we introduce two elements that increase its complexity. First, the random target placement emulates non-fixed industrial environments, promoting generalization. Second, the DRL agent receives only a 64$\times$64 RGB image from an externally-mounted camera, without proprioceptive data, forcing it to infer the robot and target poses in 3D from 2D visual inputs. This setup aims to create a versatile sim-to-real solution that is easy to implement across diverse industrial scenarios and tasks, as it does not depend on the specific robot or setup characteristics.

\subsection{Experimental setup}
\label{ES}
The pipeline for transferring DRL experience between virtual and real environments is developed using the 6-DoF ABB IRB120 industrial robotic arm~\citep{ABB_IRB120_datasheet}. The IRB120's workspace is a semicircular annular region centered at the robot's base, with inner and outer radii of \SI{25}{cm} and \SI{55}{cm}, respectively. Beyond this area, the robot is likely to encounter singularities --configurations where inverse kinematics cannot be resolved-- restricting its movement. For validation we use the UR3e from Universal Robots (UR)~\citep{UR_UR3e_datasheet}, a 6-DoF collaborative robotic arm that presents a completely different appearance, morphology, kinematics, and dynamics than the IRB120. The workspace of the UR3e is defined as a semicircular annular area centered at the robot's base, with inner and outer radii of \SI{10}{cm} and \SI{55}{cm}, similar to the IRB120. Beyond this region, the robot is prone to encounter singularities. Although the UR3e's joints provide greater freedom of movement than the IRB120, its range has been slightly limited to accommodate real-world scenario constraints.
\begin{figure}[h!]
    \centering
    \begin{subfigure}[b]{0.25\linewidth}
        \centering
        \includegraphics[width=\linewidth]{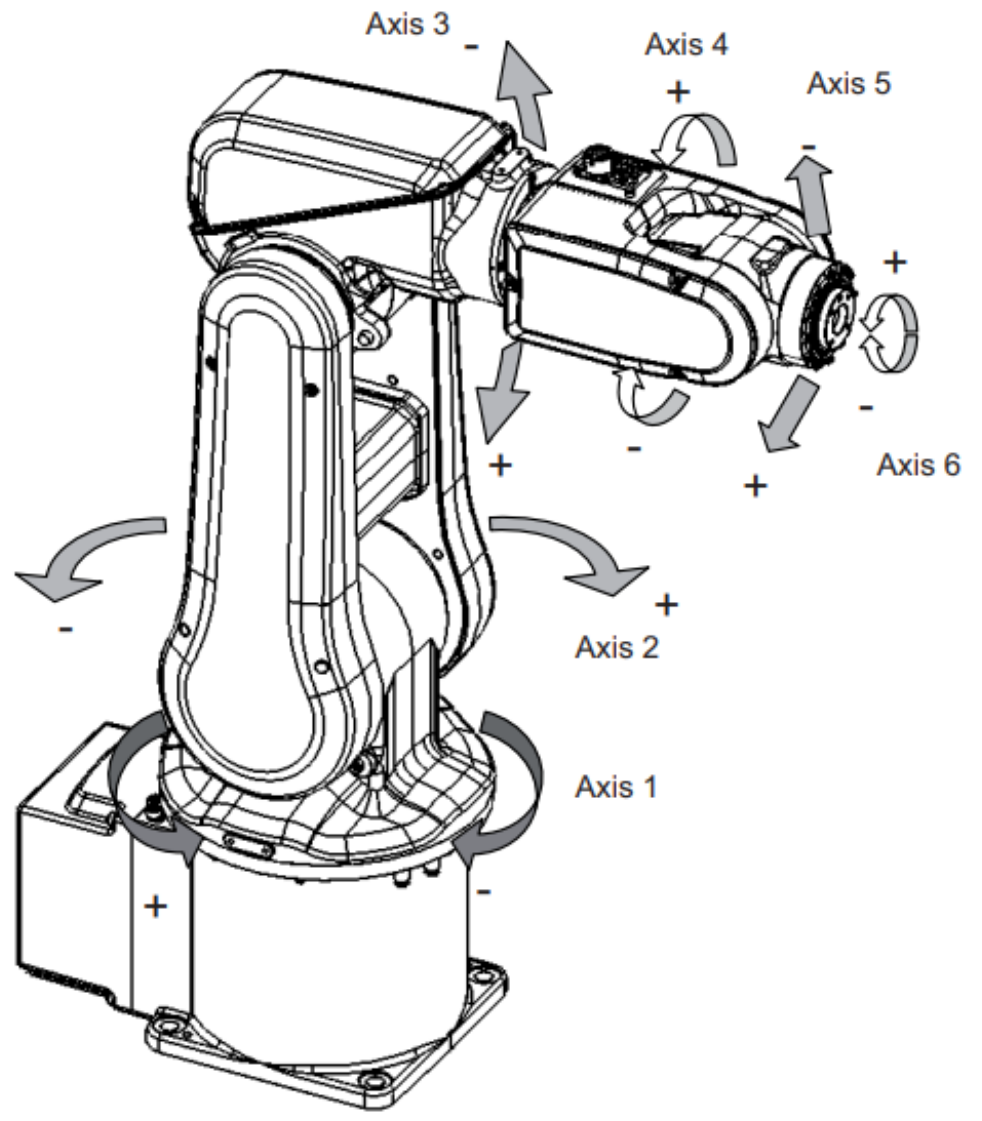}
        \caption{}
    \end{subfigure}
    \hspace{0.05\linewidth}
    \begin{subfigure}[b]{0.25\linewidth}
        \centering
        \includegraphics[width=\linewidth]{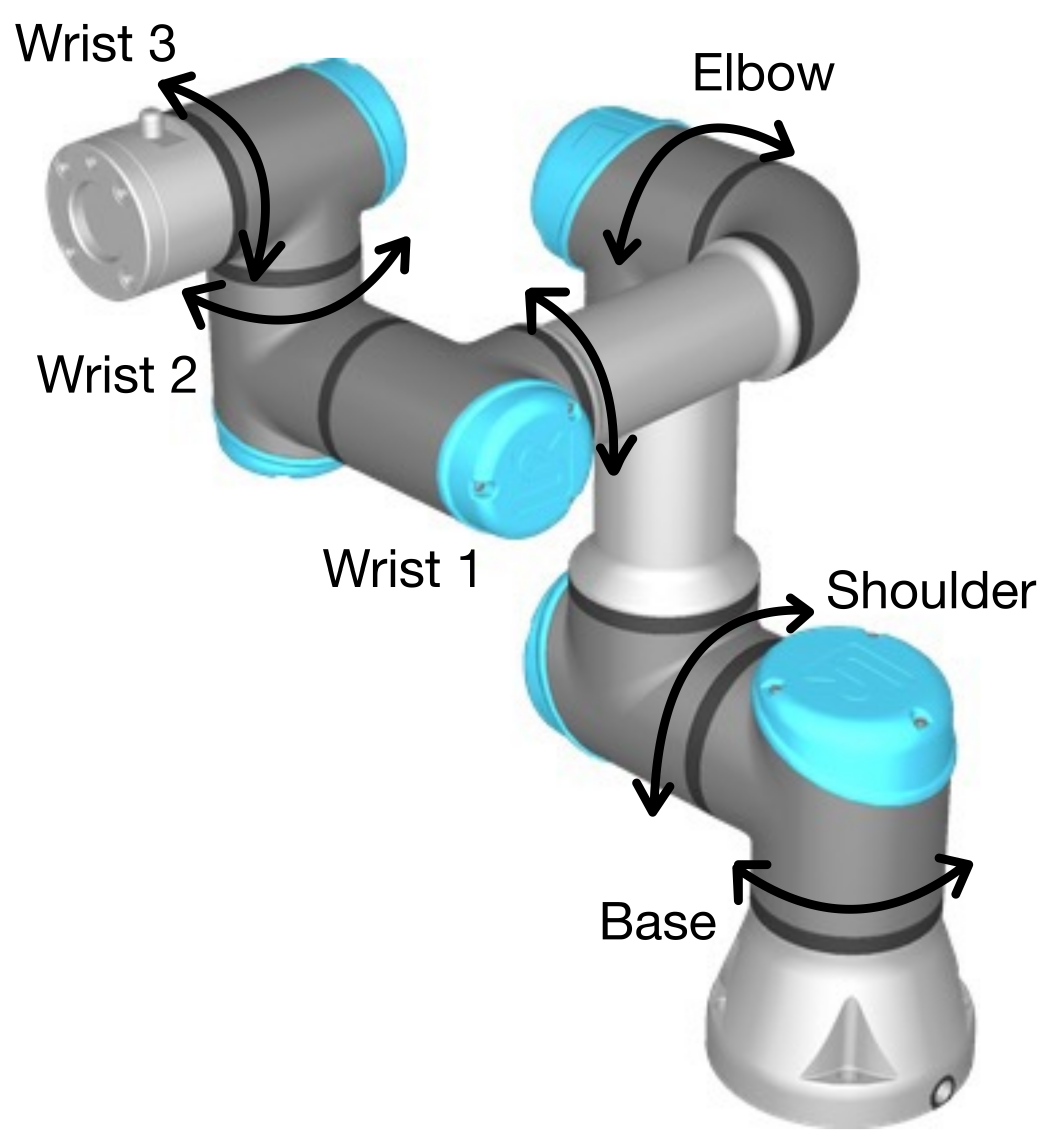}
        \caption{}
    \end{subfigure}
    \caption{IRB120 (\textit{Source:} ABB~\citep{ABB_IRB120_datasheet}) (a) and UR3e (b) axes and their rotation.}
    \label{fig:ProbDecrp_UR3eExpSetup_axes}
\end{figure}

\begin{figure}[h!]
    \centering
    \includegraphics[width=0.35\linewidth]{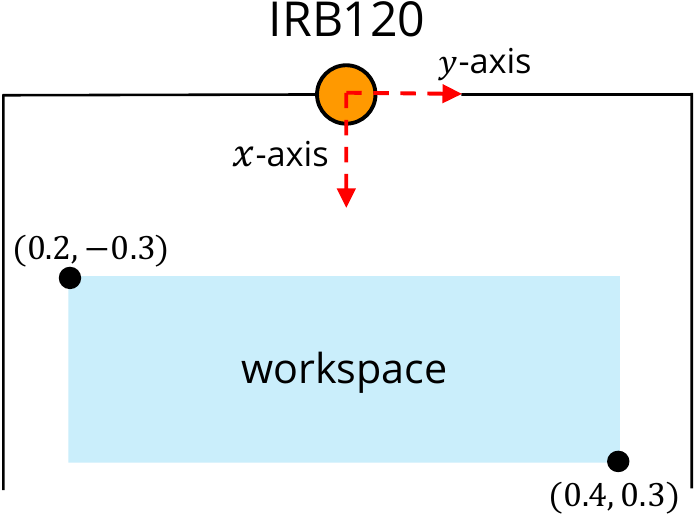}
    \caption{IRB120 workspace and axes location. The target coordinates must be within the intervals [0.2, 0.4]~m for the $x$-axis and [--0.3, 0.3]~m for the $y$-axis.}
    \label{fig:IRB120ExpSetup_workspace}
\end{figure}

\begin{figure}[h!]
    \centering
    \subfloat[]{\includegraphics[width=0.2\columnwidth]{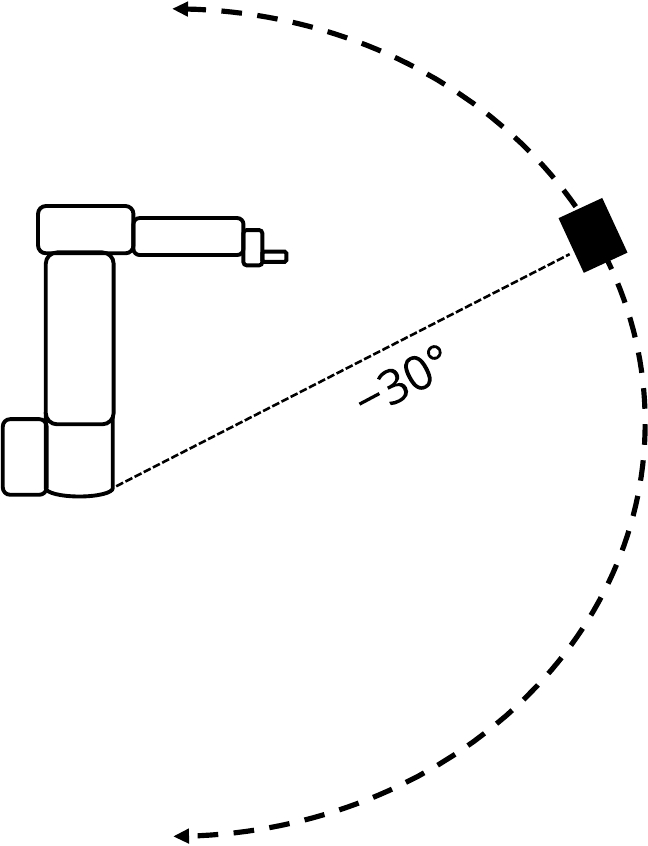}}
    \hspace{0.1\columnwidth}
    \subfloat[]{\includegraphics[width=0.35\columnwidth]{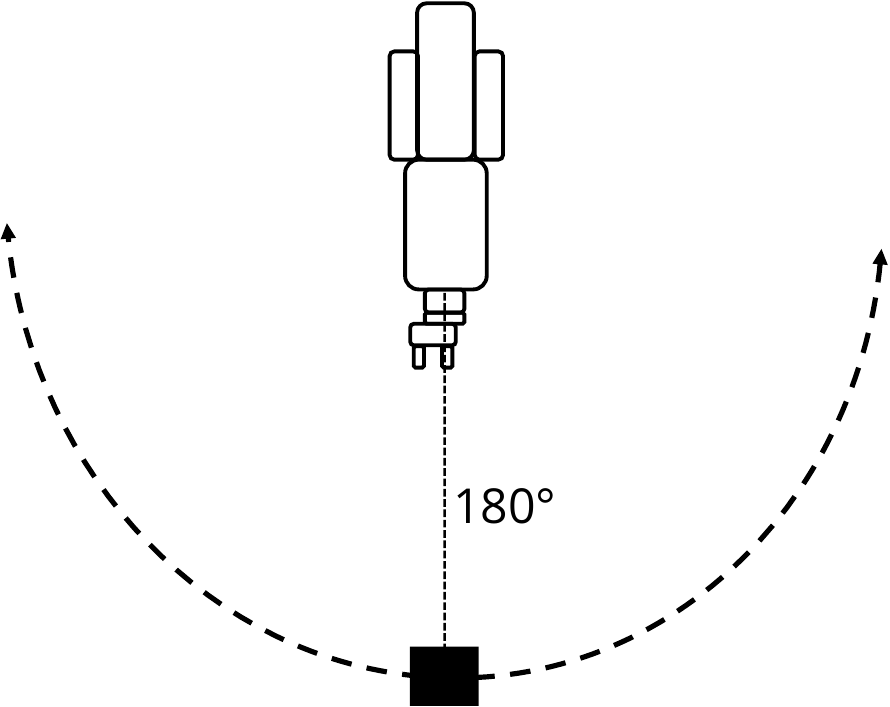}}
    \caption{Camera poses in the IRB120 setup: (a) around the $z$-axis, (b) around the $y$-axis.}
    \label{fig:IRB120ExpSetup_camerapose}
\end{figure}

\begin{figure}[h!]
    \centering
    \subfloat[]{\includegraphics[width=0.2\columnwidth]{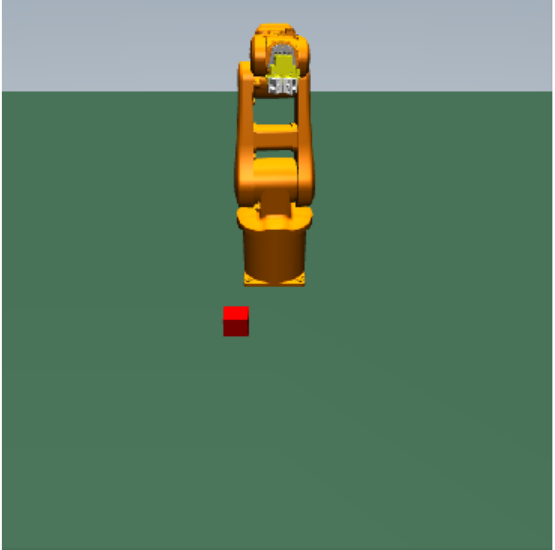}}
    \hspace{0.1\columnwidth}
    \subfloat[]{\includegraphics[width=0.2\columnwidth]{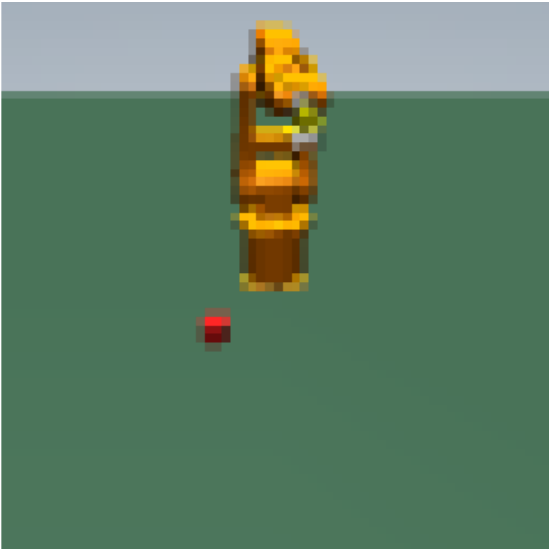}}
    \caption{IRB120 virtual observations with 640$\times$640 pixel resolution (a) and 64$\times$64 pixel resolution (b), which is the one received by the agent.}
    \label{fig:IRB120ExpSetup_setupVirtual}
\end{figure}

The \textit{virtual environments} are built using MuJoCo~\citep{Todorov2012}. The IRB120 environment features a green ground plane and a grey sky backdrop, with the IRB120 robot positioned centrally and its base on the ground. A red cube target is placed within the robot's reach, with $x$-coordinates in [0.2, 0.4]~m and $y$-coordinates in [--0.3, 0.3]~m (Figure~\ref{fig:IRB120ExpSetup_workspace}). The camera is positioned in the center of the scene, oriented \SI{180}{\degree} around the $z$-axis and tilted \SI{30}{\degree} from the horizontal on the $y$-axis, capturing the full setup and providing 2D images with depth cues for spatial interpretation (Figure~\ref{fig:IRB120ExpSetup_camerapose}). Although this setup differs from a typical industrial environment, it is simplified to focus on defining the sim-to-real transfer pipeline without added complexity, following approaches in state-of-the-art works such as~\citep{Rusu2017}. Figure~\ref{fig:IRB120ExpSetup_setupVirtual} illustrates the high-resolution IRB120 scenario and the 64$\times$64 pixel view received by the agent.

\begin{figure}[t]
    \centering
    \includegraphics[width=0.45\linewidth]{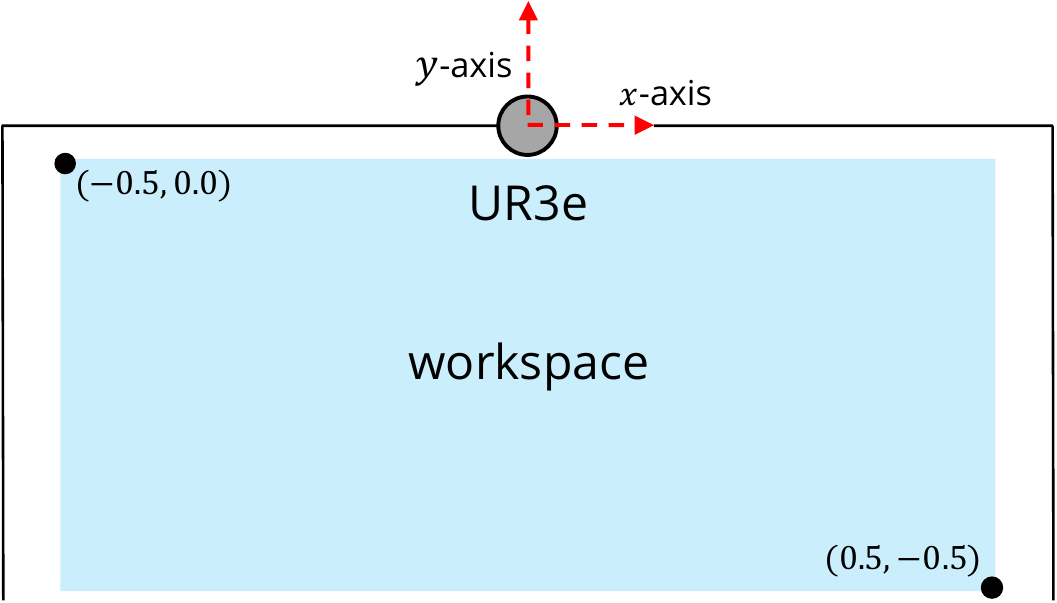}
    \caption{UR3e workspace and axes location. The target coordinates must be within the intervals [--0.5, 0.5]~m for the $x$-axis and [--0.5, 0.0]~m for the $y$-axis.}
    \label{fig:UR3eworkspace}
\end{figure}

\begin{figure}[h!]
    \centering
    \subfloat[]{\includegraphics[width=0.2\columnwidth]{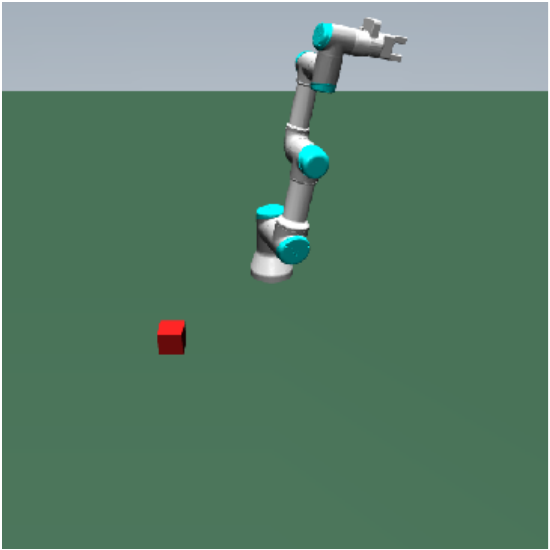}}
    \hspace{0.1\columnwidth}
    \subfloat[]{\includegraphics[width=0.2\columnwidth]{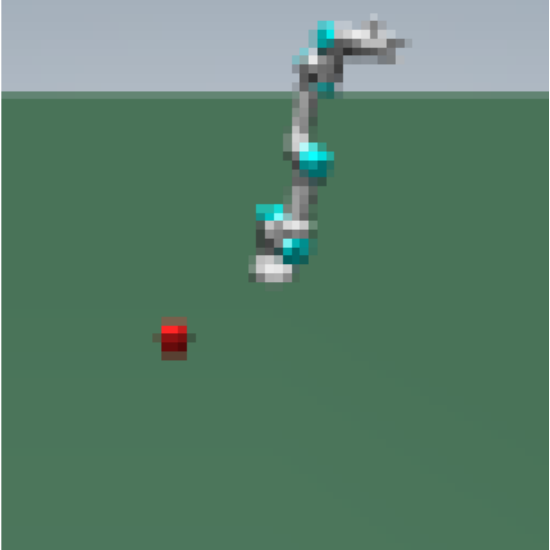}}
    \caption{UR3e virtual observations with 640$\times$640 pixel resolution (a) and 64$\times$64 pixel resolution (b), which is the one received by the agent.}
    \label{fig:UR3eExpSetup_setupVirtual}
\end{figure}

To validate the pipeline designed with the IRB120, we use the same virtual baseline environment. The target coordinates range from [--0.5, 0.5]~m along the $x$-axis and [--0.5, 0.0]~m along the $y$-axis (Figure~\ref{fig:UR3eworkspace}). In the UR3e setup, the origin of the $y$-axis is located at the robot’s base, extending into negative values as the target moves farther away. Apart from adjusting the orientation to \SI{0}{\degree} around the $z$-axis to accommodate the reference system change, the camera configuration is the same as in the IRB120 (Figure~\ref{fig:IRB120ExpSetup_camerapose}). Figure~\ref{fig:UR3eExpSetup_setupVirtual} shows the virtual environment for the UR3e and the 64$\times$64 pixel observation provided to the agent.

\begin{figure}[h!]
    \centering
    \includegraphics[width=0.65\columnwidth]{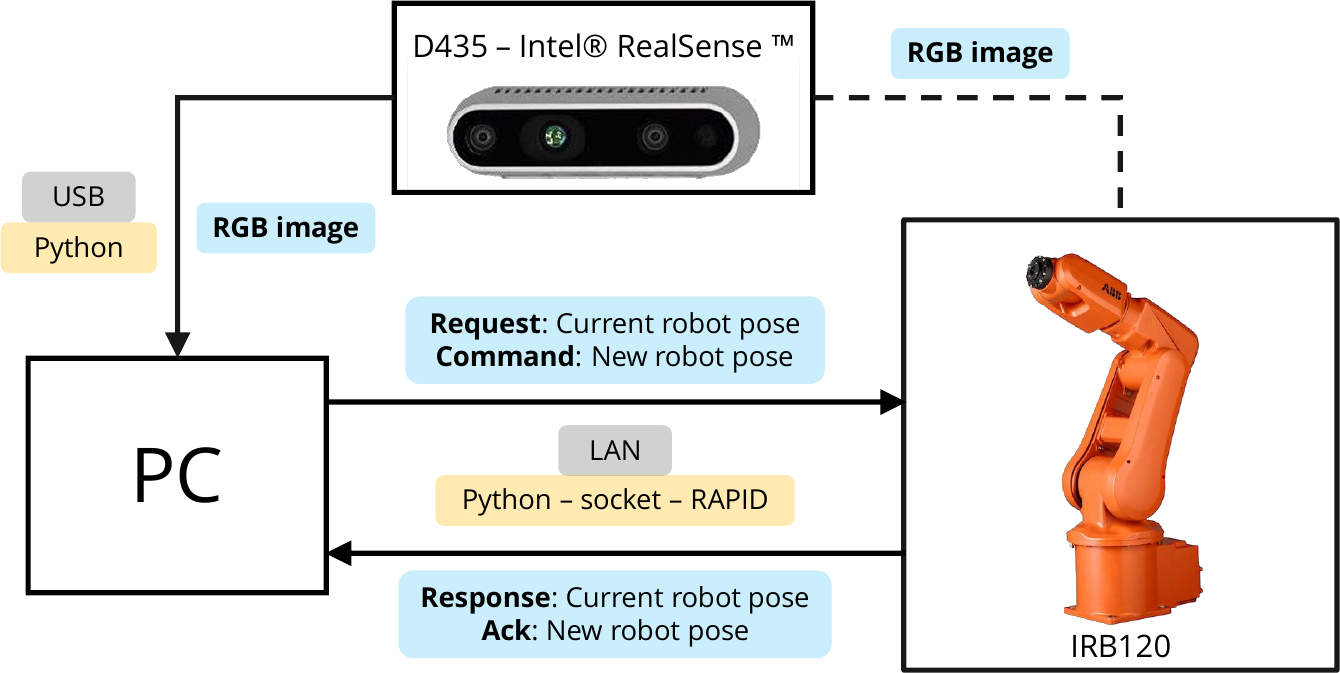}
    \caption{IRB120 real setup configuration diagram. Hardware components are connected via LAN and USB, with environment observations from the camera sent directly to the computer. Bidirectional communication between the IRB120 controller and computer is managed through sockets.}
    \label{fig:IRB120ExpSetup_communication}
\end{figure}

\begin{figure}[h!]
    \centering
    \includegraphics[width=0.65\columnwidth]{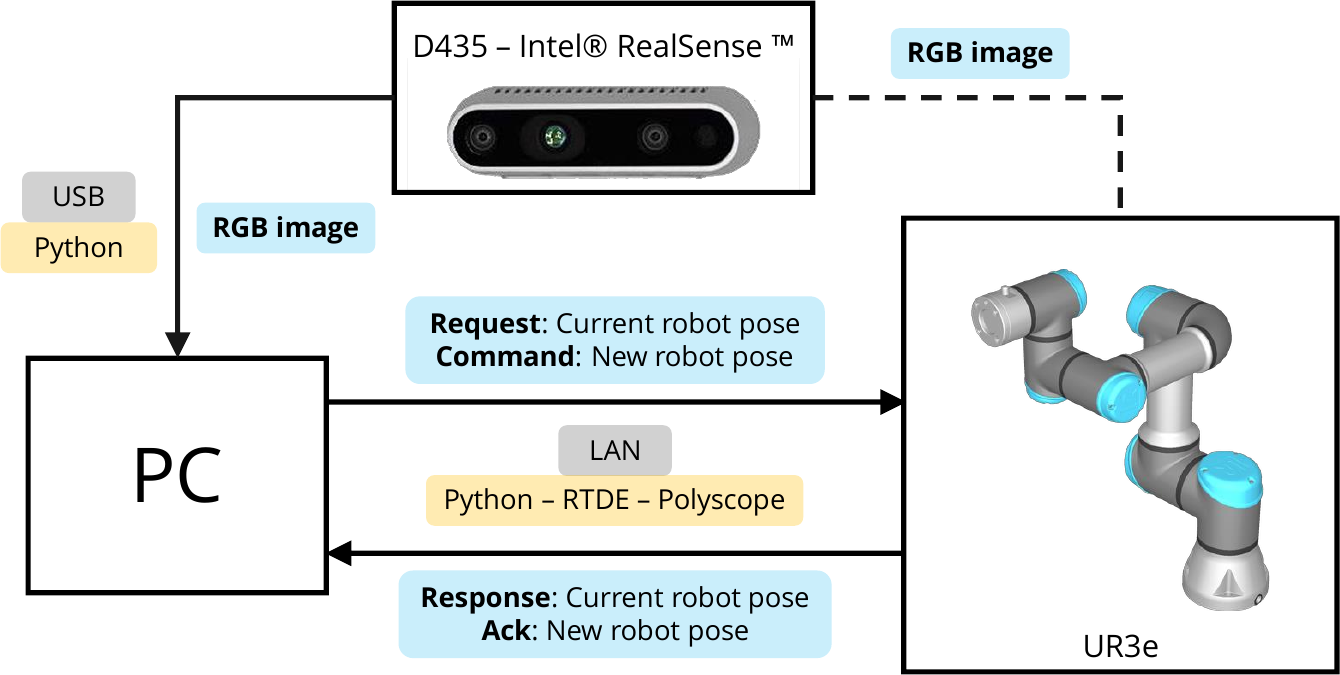}
    \caption{UR3e real setup configuration diagram. Hardware components are connected via LAN and USB, with environment observations from the camera sent directly to the computer. Bidirectional communication between the UR3e controller and computer is managed through RTDE.}
    \label{fig:UR3eExpSetup_communication}
\end{figure}

Despite efforts to replicate the virtual environment, inherent differences exist with the \textit{real environment}, including camera noise absent in simulation. Both robotic arms are centrally positioned with a metallic support, while green and grey cardboards replicate the virtual ground and sky. The Intel\textsuperscript{\textregistered}~RealSense\textsuperscript{\texttrademark}~D435 camera,\footnote{\url{https://www.intelrealsense.com/depth-camera-d435/}} equipped with depth and RGB sensors, captures the RGB input used in this setup. Communication between the camera and the computer is managed via custom drivers based on \texttt{pyrealsense2}\footnote{\url{https://github.com/IntelRealSense/librealsense}}. This camera fulfills all setup requirements, is affordable (around 350\euro), and supports future depth-based extensions. The IRB120 is connected to the computer via LAN and controlled through Python-based scripts and RAPID programs on the robot. Commands from the agent to the robot include joint orientation requests, gripper coordinates, and updated joint positions, while the robot responds with current positions and confirmations. Figure~\ref{fig:IRB120ExpSetup_communication} illustrates the configuration and communication structure. The IRB120 computer setup includes an Intel\textsuperscript{\textregistered}~Core\textsuperscript{\texttrademark}~i7-12700 CPU, \SI{128}{GB} DDR5 RAM, an NVIDIA RTX 3070 GPU, and a \SI{1}{TB} SSD, running Ubuntu 20.04 LTS, Python 3.8, and PyTorch v1.13.1~\citep{Stevens2020}. Conversely, the UR3e is connected to the computer through a LAN connection using the Real-Time Data Exchange (RTDE) protocol. Communication between the robot and the computer is managed by structured Python methods that handle data exchange. The UR3e's programming is implemented using Polyscope instructions, while the computer setup includes Ubuntu 20.04 LTS, Python 3.8, and PyTorch v1.13.1~\citep{Stevens2020} on an Intel\textsuperscript{\textregistered}~Core\textsuperscript{\texttrademark}~i7-3770 CPU with \SI{8}{GB} RAM. Figure~\ref{fig:UR3eExpSetup_communication} shows the configuration and communication structure.

\begin{figure}[h!]
    \centering
    \subfloat[]{\includegraphics[width=0.15\columnwidth]{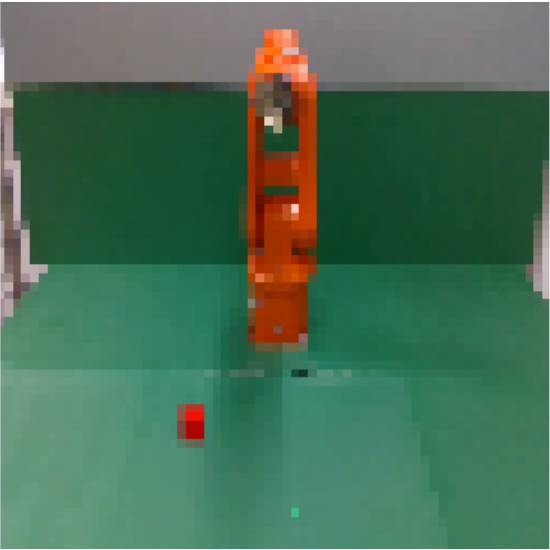}}
    \hspace{0.03\columnwidth}
    \subfloat[]{\includegraphics[width=0.15\columnwidth]{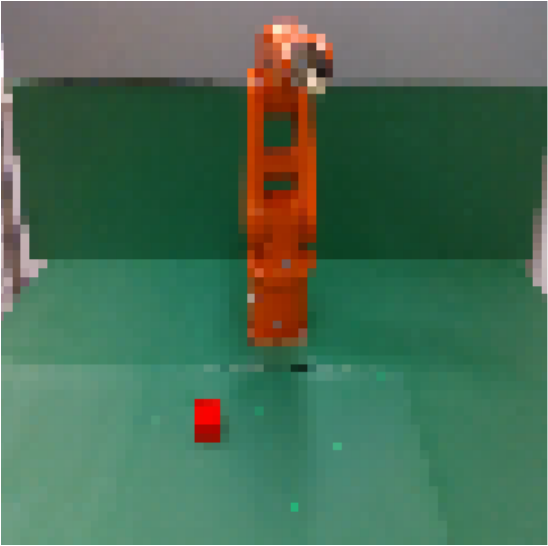}}
    \hspace{0.03\columnwidth}
    \subfloat[]{\includegraphics[width=0.15\columnwidth]{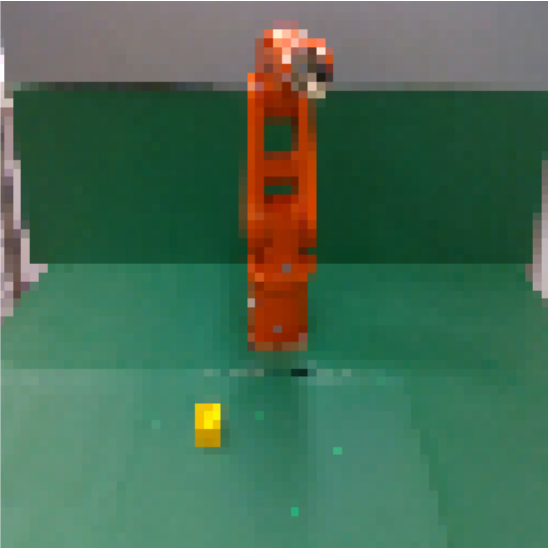}}
    \hspace{0.03\columnwidth}
    \subfloat[]{\includegraphics[width=0.15\columnwidth]{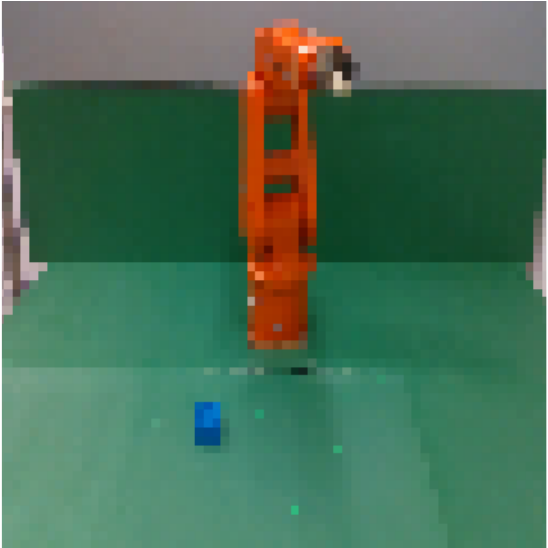}}
    \hspace{0.03\columnwidth}
    \subfloat[]{\includegraphics[width=0.15\columnwidth]{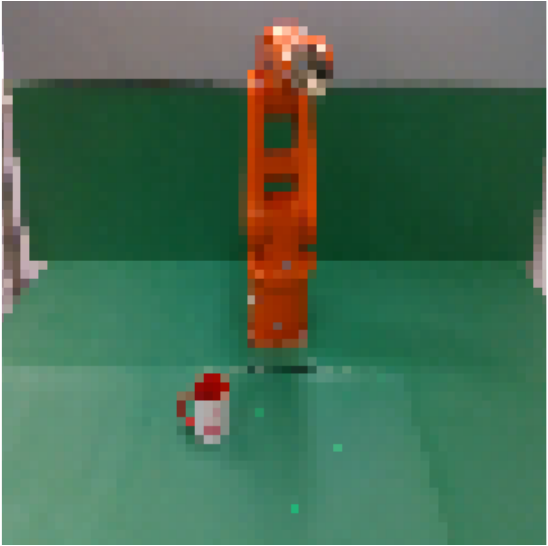}}
    \caption{IRB120 64$\times$64 pixel observations in the real setup with the AR target (a), the red (b), yellow (c), and blue (d) LEGO\textsuperscript{\textregistered}~cubes, and the white and red mug (e).}
    \label{fig:IRB120ExpSetup_setupReal}
\end{figure}

\begin{figure}[t]
    \centering
    \subfloat[]{\includegraphics[width=0.15\columnwidth]{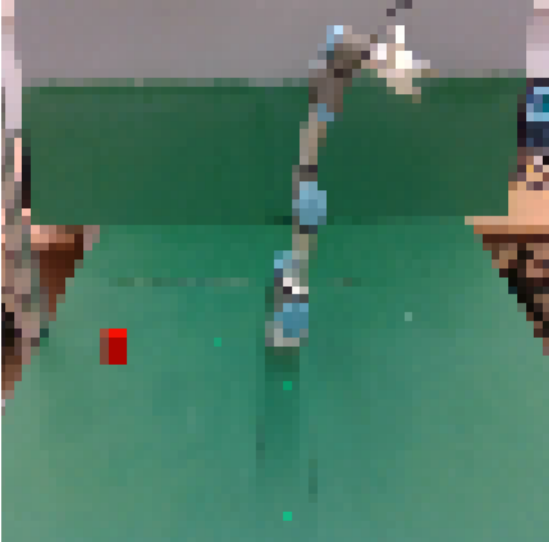}}
    \hspace{0.03\columnwidth}
    \subfloat[]{\includegraphics[width=0.15\columnwidth]{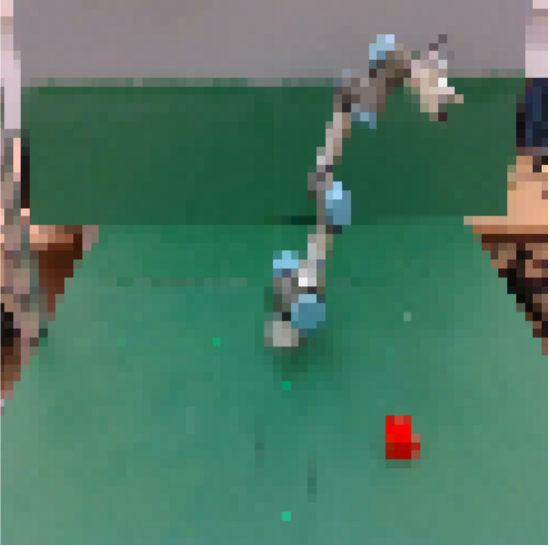}}
    \hspace{0.03\columnwidth}
    \subfloat[]{\includegraphics[width=0.15\columnwidth]{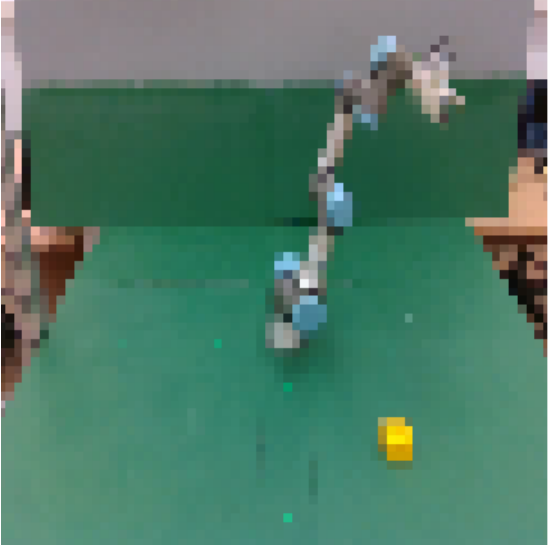}}
    \hspace{0.03\columnwidth}
    \subfloat[]{\includegraphics[width=0.15\columnwidth]{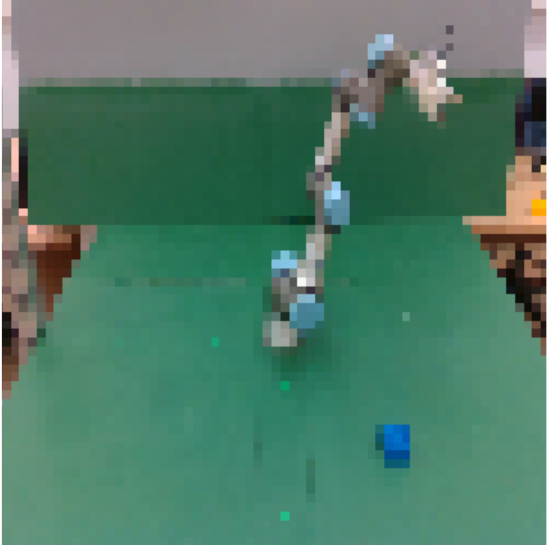}}
    \hspace{0.03\columnwidth}
    \subfloat[]{\includegraphics[width=0.15\columnwidth]{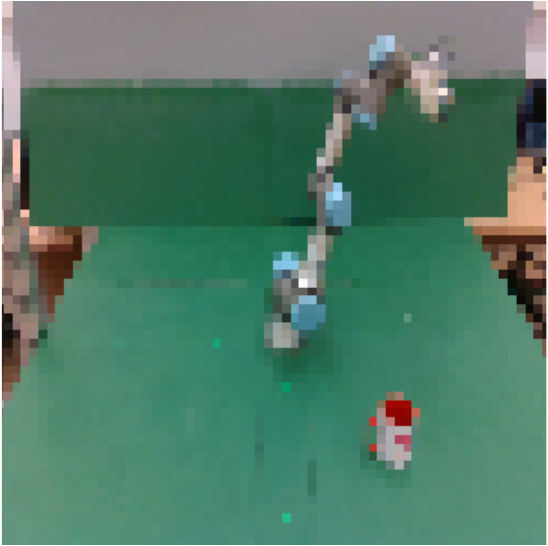}}
    \caption{UR3e 64$\times$64 pixel observations in the real setup with the AR target (a), the red (b), yellow (c), and blue (d) LEGO\textsuperscript{\textregistered}~cubes, and the white and red mug (e).}
    \label{fig:UR3eExpSetup_setupReal}
\end{figure}

To facilitate the post-training evaluation in the real setup, target objects are projected within the workspace using augmented reality (AR), speeding up the process and eliminating the need for manual placement. Target positions are pre-defined using 125 ArUco markers~\citep{Garrido2014}, enabling fast, accurate rendering. After the initial zero-shot evaluation, the agents are further validated with real objects, including a red, yellow, and blue LEGO\textsuperscript{\textregistered}~cube, and a red and white mug (Figure~\ref{fig:IRB120ExpSetup_setupReal} and Figure~\ref{fig:UR3eExpSetup_setupReal}).

\FloatBarrier
\section{Method}
\label{Method}
\begin{figure}[h!]
    \centering
    \includegraphics[width=0.8\columnwidth]{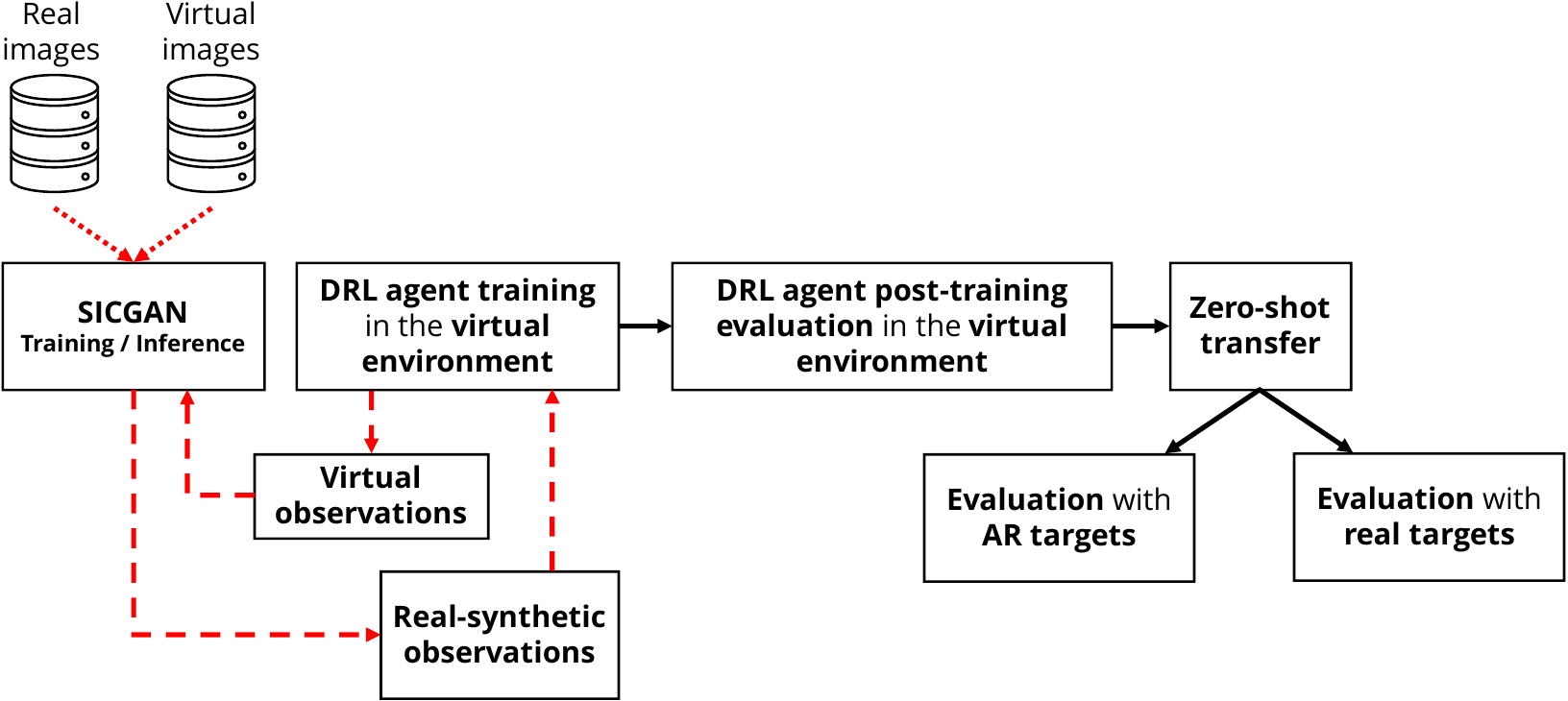}
    \caption{Zero-shot transfer pipeline using SICGAN. The DRL agent is trained on real-synthetic observations generated by the SICGAN and evaluated post-training in the virtual environment. Once verified, the agent is directly deployed in the real environment (zero-shot transfer), where it is tested with AR and real targets. Red dashed arrows represent the training phase, while real-world execution is represented by solid black arrows.}
    \label{fig:methodZS}
\end{figure}

Figure~\ref{fig:methodZS} illustrates the pipeline used to achieve a zero-shot transfer with the DA approach based on our SICGAN implementation. Initially, the SICGAN is trained \emph{off-process} on a labeled dataset containing both virtual and real observations. Given the industrial application requirements, we opted to perform sim-to-real translation from virtual to real images, enabling agent training in a resource-rich virtual environment. This approach allows us to leverage GPU-based training during simulation while avoiding such expensive hardware in the industrial setting, where the SICGAN is no longer needed. Following virtual training on real-synthetic images, which are obtained using the GAN in inference mode, the virtually-trained agent is deployed directly in the real environment, using raw camera inputs for evaluation. This way, the proposed solution is fully scalable to other robotic manipulators in industrial environments. Furthermore, the approach eliminates the need for intensive computational resources, as the agent operates in inference mode during zero-shot deployment, directly receiving the camera image as input. Performance is assessed with both ArUco AR targets and real targets to validate the agent's generalization capability.

\subsection{StyleID-CycleGAN (SICGAN)}
\label{Method_SICGAN}
To mitigate common artifacts (e.g., color bleeding or distorted edges) present in CycleGAN-based architectures, SICGAN introduces two key enhancements inspired by recent advances in generative modeling. First, SICGAN replaces batch normalization with demodulated convolutions, following the approach introduced in StyleGAN~\citep{Karras2018} and StyleGANv2~\citep{Karras2019}. In this setting, convolutional weights are first modulated based on input-dependent style or feature vectors, allowing the network to adapt filter responses to specific image attributes. To ensure stability, the modulated weights are then demodulated by normalizing them channel-wise, preventing uncontrolled amplification or suppression of signal magnitudes. In our implementation, the scaling and standarization is carried out per feature map using its mean and standard deviation. This combination leads to more consistent feature transformations and effectively reduces the presence of artifacts during image translation.

Secondly, SICGAN incorporates an \emph{identity loss}~\eqref{eq:CycleGAN_IdentLoss}, which encourages the generator to preserve key image features when processing samples from the target domain. By penalizing unnecessary changes in such cases, identity loss promotes semantic and color consistency, further improving the fidelity of translated images, and preserving original image features when target domain images are reprocessed by their generator.
\begin{equation}
\resizebox{.7\hsize}{!}{$
\begin{aligned}
\mathcal{L}_{\mathrm{id}} &= \mathcal{L}_{\mathrm{id}}^{G} + \mathcal{L}_{\mathrm{id}}^{F} \\
&= \mathbb{E}_{y \sim p_{\mathcal{Y}}} \left[ \| G(y) - y \|_1 \right] + \mathbb{E}_{x \sim p_{\mathcal{X}}} \left[ \| F(x) - x \|_1 \right]
\end{aligned}$}
\label{eq:CycleGAN_IdentLoss}
\end{equation}
Equation~\eqref{eq:CycleGAN_TotLossSICGAN} defines our final generators' loss function.
\begin{equation}
\resizebox{.8\hsize}{!}{$
\begin{aligned}
\mathcal{L}(G, F, D_{\mathcal{X}}, D_{\mathcal{Y}}) &= \mathcal{L}_{\mathrm{GAN}}(G, D_{\mathcal{Y}}, \mathcal{X}, \mathcal{Y}) + \mathcal{L}_{\mathrm{GAN}}(F, D_{\mathcal{X}}, \mathcal{Y}, \mathcal{X}) \\
&\quad + \lambda_{\mathrm{cyc}} \left( \mathcal{L}_{\mathrm{cyc}}^{G \rightarrow F} + \mathcal{L}_{\mathrm{cyc}}^{F \rightarrow G} \right) \\
&\quad + \lambda_{\mathrm{id}} \left( \mathcal{L}_{\mathrm{id}}^{G} + \mathcal{L}_{\mathrm{id}}^{F} \right)
\end{aligned}$}
\label{eq:CycleGAN_TotLossSICGAN}
\end{equation}
where $\mathcal{L}_{\mathrm{GAN}}(G, D_{\mathcal{Y}}, \mathcal{X}, \mathcal{Y})$ is the adversarial loss for the generator $G$, which encourages it to translate samples from $\mathcal{X}$ into indistinguishable outputs from real samples in domain $\mathcal{Y}$, as judged by discriminator $D_{\mathcal{Y}}$. Similarly, $\mathcal{L}_{\mathrm{GAN}}(F, D_{\mathcal{X}}, \mathcal{Y}, \mathcal{X})$ is the adversarial loss for the inverse mapping $F$, which ensures that images translated from $\mathcal{Y}$ to $\mathcal{X}$ are realistic with respect to $D_{\mathcal{X}}$. The $\lambda$ hyperparameters are set to $\lambda_{\mathrm{cyc}} = 10$ and $\lambda_{\mathrm{id}} = 0.1$.

\begin{figure}[h!]
    \centering
    \includegraphics[width=\columnwidth]{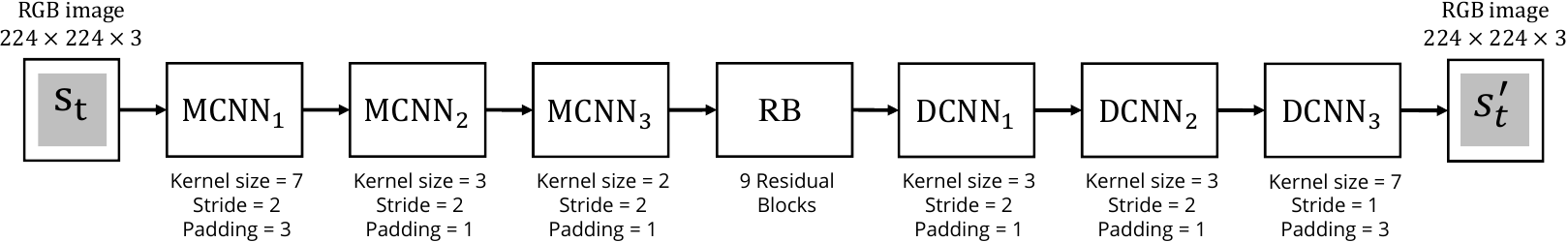}
    \caption{SICGAN generator architecture. The input, a 224$\times$224 RGB image, is processed through a series of modulated CNN layers (MCNN), followed by nine residual blocks (RB), and concludes with demodulated CNN (DCNN) layers, producing an output of the same 224$\times$224 RGB resolution.}
    \label{fig:sicganGen}
\end{figure}

\begin{figure}[h!]
    \centering
    \includegraphics[width=0.7\columnwidth]{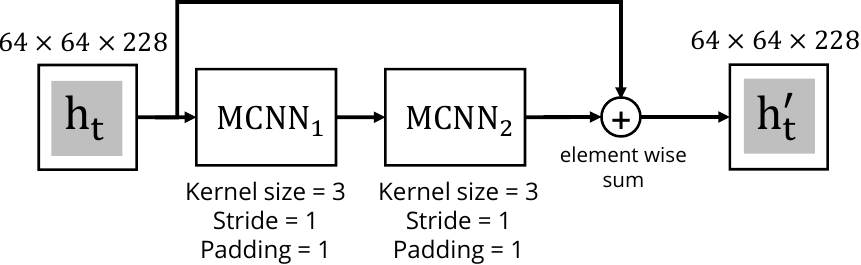}
    \caption{SICGAN residual block architecture. It consists of two modulated CNN layers (MCNN), with their output added element-wise to the input.}
    \label{fig:sicganRB}
\end{figure}

\begin{figure}[h!]
    \centering
    \includegraphics[width=0.7\columnwidth]{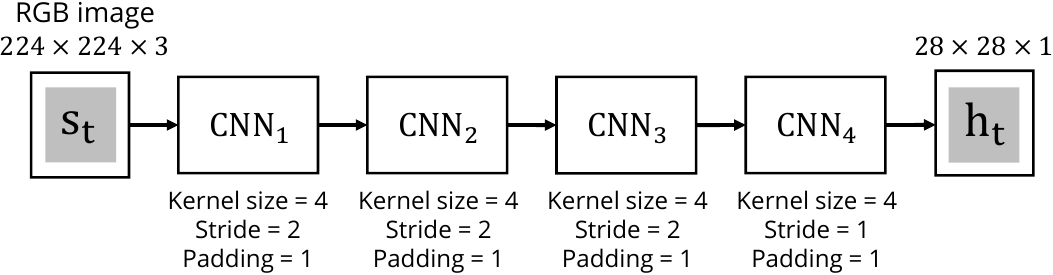}
    \caption{SICGAN discriminator architecture. The input, a 224$\times$224 RGB image, passes through a series of CNN layers, producing a 28$\times$28 prediction matrix as output.}
    \label{fig:sicganDis}
\end{figure}

The SICGAN generator is based on ResNet~\citep{He2015} and uses residual blocks to maintain key features during translation while preventing vanishing gradients. Figure~\ref{fig:sicganGen} and Figure~\ref{fig:sicganRB} illustrate the generator and residual block architectures, respectively. The input is a 224$\times$224 RGB image, processed in a classic autoencoder layout: an initial demodulated convolution stage, nine residual blocks, and an upsampling phase using transposed convolutions to restore the input image size. The intermediate activation function is ReLU, and the final layer uses hyperbolic tangent (\texttt{tanh}), producing pixel values in [--1, 1]. Figure~\ref{fig:sicganDis} shows the architecture of the discriminator. It processes a 224$\times$224 RGB image through successive CNN layers~\citep{Lecun2015, Goodfellow2016, Li2022_b}, concluding with a PatchGAN-based output~\citep{Li2016, Isola2017} that generates a 28$\times$28 prediction matrix. PatchGAN discriminators assess individual patches within the image rather than the whole capture, with each matrix element indicating whether a specific region appears real or synthetic. This localized approach enables the discriminator to focus on fine-grained features, improving the generator's training outcome. Leaky ReLU with a 0.2 negative slope is used in intermediate layers to stabilize gradients by allowing small negative values.

The dataset for training each SICGAN, one for the IRB120 and another for the UR3e, consists of 1,300 labeled 224$\times$224 RGB images from the virtual and real environments, capturing random robot poses without the target and split into a 70/30 train-test ratio. The adversarial loss uses Mean Squared Error (squared L2 norm), while cycle consistency and identity losses are based on the Mean Absolute Error (MAE) or L1 norm. The Adam optimizer is applied with $\beta_{1} = 0.5$ and $\beta_{2} = 0.999$, and weights are initialized from a Normal distribution $w \sim N(0, 0.02^{2})$. Key hyperparameters are listed in Table~\ref{tab:sicganHyperparam}. The SICGAN training process lasted 27 hours on a PC running Ubuntu 20.04, Python 3.8, and PyTorch v1.13.1~\citep{Stevens2020} equipped with an Intel\textsuperscript{\textregistered}~Core\textsuperscript{\texttrademark}~i9-10900KF CPU at \SI{3.70}{GHz}, \SI{64}{GB} of DDR4 RAM, an NVIDIA GeForce RTX 2080 Ti GPU, and a \SI{2}{TB} M.2 SSD. The optimal SICGAN model was selected by identifying those with the best balance between the training and validation losses, complemented by a visual assessment of the translation quality in both directions (i.e., sim-to-real and real-to-sim).
\begin{table}[h!]
    \centering
    \caption{SICGAN model hyperparameters.}
    \label{tab:sicganHyperparam}
    \renewcommand{\arraystretch}{1.0}
    \resizebox{0.6\columnwidth}{!}{%
    \begin{tabular}{>{\centering\arraybackslash}m{4.5cm} >{\centering\arraybackslash}m{8cm}}
        \toprule
        \textbf{Hyperparameter} & \textbf{Value} \\ \midrule[2pt]
        Number of epochs & 500 (max) \\
        Batch size & 1 \\ 
        Learning rate & $5e^{-4}$ \\
        Generator optimizer & Adam ($\beta_{1} = 0.5$, $\beta_{2} = 0.999$) \\ 
        Discriminator optimizer & Adam ($\beta_{1} = 0.5$, $\beta_{2} = 0.999$) \\ 
        Loss functions & Squared L2 norm (adversarial) \linebreak L1 loss (cycle consistency and identity) \\ 
        Weight initialization & $w \sim N(0, 0.02^{2})$ \\ 
        Image size & 224$\times$224$\times$3 \\ \bottomrule
    \end{tabular}%
    }
\end{table}

As part of our work to validate the effectiveness of the proposed SICGAN and highlight its advantages over existing approaches, we compare its performance against two reference models: the original CycleGAN and a more advanced vision-based architecture leveraging visual transformers, UVCGANv2, which was introduced in Section~\ref{RW}. In both cases, the models are trained using our IRB120 dataset, and then we evaluate their ability to translate observations from the virtual to the real domain. This evaluation involves analyzing the visual quality of the translated images and quantitatively comparing their RGB histograms with those of real images. Additionally, the Wasserstein Distance (WD)~\citep{Rubner2000} is computed for each RGB channel to objectively assess the similarity between the translated and real distributions.

\subsection{DRL agent}
\label{Method_DRLagent}
One of the most well-suited DRL algorithms is the Asynchronous Advantage Actor-Critic (A3C)~\citep{Mnih2016} due to its variance reduction via error estimators and efficient parallelization, enabling stable learning and faster convergence. The A3C framework consists of an \emph{actor} that optimizes the policy $\pi(s)$ and a \emph{critic} that estimates the state-value function $V(s)$. A3C differs from its synchronous counterpart, Advantage Actor-Critic (A2C), by using asynchronous updates, allowing multiple agents to learn in parallel on separate environment instances. Each agent locally computes gradients and asynchronously updates a shared global network, enhancing efficiency, reducing experience correlation, and improving performance.

\begin{figure}[h!]
    \centering
    \includegraphics[width=0.7\columnwidth]{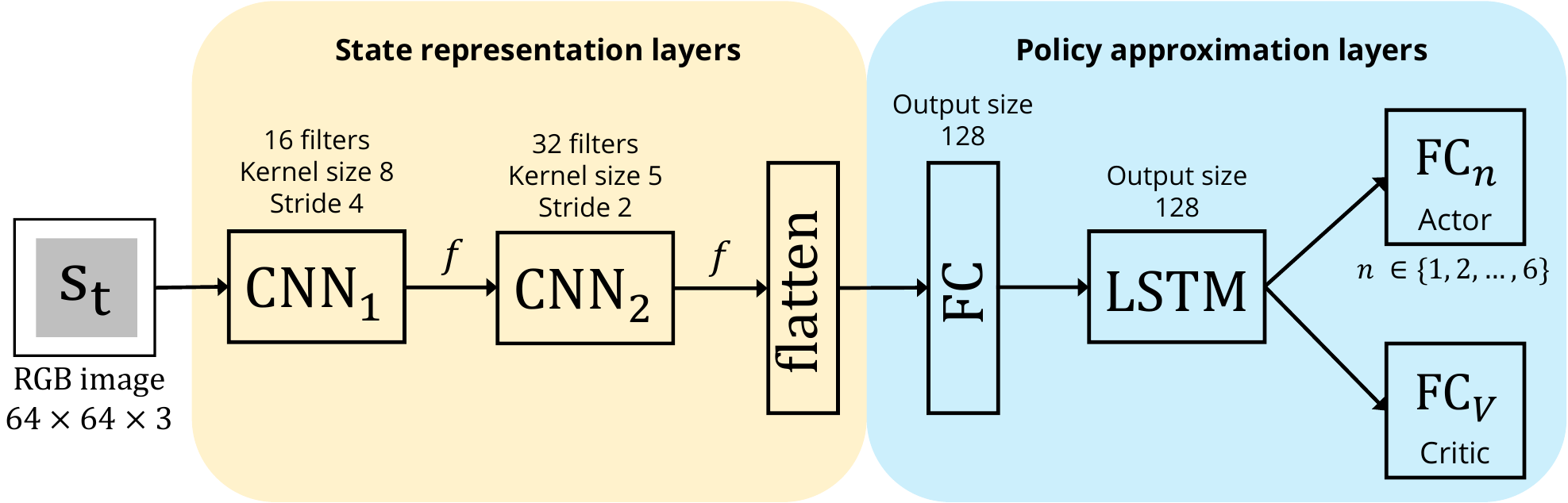}
    \caption{Implemented A3C architecture. The input is a 64$\times$64 RGB image processed by CNN layers with 16 and 32 filters to learn the state representation. Activations pass through FC and LSTM layers for temporal information, producing joint policies via softmax and estimating the state value. The ReLU function ($f$) is the activation function.}
    \label{fig:A3Carch}
\end{figure}

Figure~\ref{fig:A3Carch} shows the model architecture. The 64$\times$64 RGB image is processed through two CNN layers with 16 filters of size 8$\times$8 (stride 4) and 32 filters of size 5$\times$5 (stride 2), respectively, to extract key image features. Each CNN layer applies a ReLU non-linearity, $f(x) = \text{max}(0, x)$. The output is flattened and passed through a FC layer. Following the FC layer, a LSTM with 128 hidden states captures the sequential nature of the task, storing information across steps. The output then branches into six FC heads for the A3C actor (one per robot joint) and one for the A3C critic to estimate $V(s)$. A softmax function translates the actor outputs into joint action probabilities, forming the policy $\pi_{n}$ for each joint based on the Markov Decision Process (MDP) action definitions. Although robot joints accept continuous commands, a discrete action space is designed to accommodate A3C's discrete operation.

In the MDP framework, defined by states $S$, actions $A$, and rewards $R$, we use a 64$\times$64 pixel observation as the state and design an appropriate action set and reward function. We selected a discrete action space that commands the joints orientation, suitable for the A3C algorithm and efficient for robotic control tasks, as studies indicate that finer discretization or continuous actions can increase computational demands without necessarily improving outcomes~\citep{Kanervisto2020, Tang_2020, Pan2022}.

For rewards, while sparse configurations are ideal to avoid guiding the agent too explicitly, they often hinder learning by providing limited feedback. Instead, we implemented a reward structure that provides feedback on the agent's distance from the target at each step, with greater penalties for farther positions and compensatory rewards upon reaching the goal. We also include variable rewards based on the number of steps taken, incentivizing efficiency. Table~\ref{tab:MDPsetup} summarizes the selected MDP based on its results when trained using the A3C algorithm in the virtual environment. The Maximum Position Increment (MPI) is obtained by multiplying the joint range by 0.015. For more details on the different MDPs designed and tested, refer to~\citep{Guitta2022}.
\begin{table}[h!]
    \centering
    \caption{Definition of the MDP.}
    \label{tab:MDPsetup}
    \renewcommand{\arraystretch}{1.0}
    \begin{threeparttable}
    \resizebox{0.7\columnwidth}{!}{
        \begin{tabular}{>{\centering\arraybackslash}m{1.5cm} >{\centering\arraybackslash}m{3cm} >{\centering\arraybackslash}m{5cm} >{\centering\arraybackslash}m{3.5cm}}
            \toprule
            \textbf{Model} & \textbf{Action space}\tnote{a} & \textbf{Positive reward if target reached}\tnote{b} & \textbf{Negative reward per step}\tnote{c} \\
            \midrule
            \texttt{M2} & $0$ \linebreak $\pm \text{MPI}$ \linebreak $\pm \text{MPI}/10$ \linebreak $\pm \text{MPI}/100$ & $+70$ & $-(2 \cdot dist)^2$ \\
            \bottomrule
        \end{tabular}
    }
    \begin{tablenotes}
    	\scriptsize
        \item[a] MPI means Maximum Position Increment and \texttt{reward\_dist} rewarding distance.
        \item[b] \texttt{ep\_length} means episode length.
        \item[c] $dist$ is the distance between the gripper and the target.
    \end{tablenotes}
    \end{threeparttable}
\end{table}

\subsubsection{DRL agent training and evaluation procedures in the virtual environment}
\label{Method_DRLagent_trainingEvalVEnv}
During the agent training process, the raw observation from the virtual environment is translated into a real-like image using the SICGAN, excluding the target. The target is added afterward, with its position randomly selected from pre-saved ArUco locations. This provides the DRL agent with a complete real-synthetic observation while still acting in the virtual environment, thus preserving the cost and time efficiency of virtual learning and allowing for the parallelized benefits of the A3C architecture. A rewarding distance of \SI{5}{cm} is used, with a post-training evaluation of \SI{10}{cm}. At the start of each episode, the robot's initial joint angles and the target’s position are randomly sampled. Joint angles follow a uniform distribution within $\pm 15\%$ of their working range, while the target's $x$ and $y$ coordinates are uniformly sampled from $U(0.2, 0.4)$~m and $U(-0.3, 0.3)$~m, respectively, creating a target area of 20x60cm.

The agent was trained for 35M steps with interim evaluations every 50k steps using 40 fixed robot and target positions for fair comparison. Weights are initialized orthogonally. Table~\ref{tab:TrainingEvalHyperparams} gathers the hyperparameters, selected based on recommendations from~\citep{Rusu2017} and standard values effective in similar settings. No sensitivity analysis was conducted, as results were satisfactory; future work could further optimize these parameters.
\begin{table}[h!]
\centering
\caption{A3C agent training hyperparameters.}
\label{tab:TrainingEvalHyperparams}
\renewcommand*{\arraystretch}{1.0}
\resizebox{0.6\columnwidth}{!}{%
\begin{tabular}{>{\centering\arraybackslash}m{5cm} >{\centering\arraybackslash}m{7cm}}
    \toprule
    \textbf{Hyperparameter} & \textbf{Value} \\
    \midrule[2pt]
    Seed & 123 (training) \linebreak 803 (post-training evaluation) \\
    Training steps & 35~M \\
    Episode length & 50 steps or target reached \\
    Success distance & \SI{5}{cm} (training) \linebreak \SI{10}{cm} (post-training evaluation) \\
    Evaluation interval & 50~k steps \\
    Evaluation episodes & 40 episodes (training) \linebreak 1,000 (post-training evaluation) \\
    Discount factor ($\gamma$) & 0.99 \\
    RMSProp learning rate & 1$e^{-4}$ \\
    RMSprop decay & 0.99 \\
    Entropy weight ($\beta$) & 0.01 \\
    Trace decay factor ($\lambda$) & 1 \\
    \bottomrule
\end{tabular}%
}
\end{table}

Post-training evaluation includes 1,000 episodes with random initial configurations and a new seed for unbiased comparison. Performance metrics include accuracy, defined in~\eqref{eq:Accuracy} as the percentage of episodes in which the agent reaches the target within \SI{10}{cm}, along the mean episode length, episode returns, and failure distances.
\begin{equation}
    \mathrm{Accuracy}~[\%] = \frac{1}{N} \sum_{i=1}^{N} \mathbb{I}(d_i \leq 10\,\mathrm{cm}) \times 100
    \label{eq:Accuracy}
\end{equation}
\FloatBarrier
\subsubsection{DRL agent evaluation procedure in the real environment}
\label{Method_DRLagent_trainingEvalREnv}
Following the agent's evaluation in the virtual environment, a zero-shot transfer is conducted in the real environment directly using the observations captured by the camera (i.e., the SICGAN is not used in the zero-shot deployment). Each ArUco position is tested five times with random initial robot poses, producing a heat map of the agent's accuracy across positions. To complete the workspace map, values are interpolated using the nearest neighbors. After testing with AR targets, real objects (i.e., red, yellow, and blue LEGO\textsuperscript{\textregistered} cubes, and a red and white mug, as shown in Figure~\ref{fig:IRB120ExpSetup_setupReal} and Figure~\ref{fig:UR3eExpSetup_setupReal}) are used to assess generalization. For the IRB120 setup, we evaluated the real red LEGO\textsuperscript{\textregistered} cube in ten new positions, from which four successful positions were selected for further testing with objects of varying colors and shapes. In the case of the UR3e, as we are verifying the robustness of the pipeline, the evaluation with real targets is done directly in five unseen workspace positions.

\section{Results and discussion}
\label{Results_Discussion}
\subsection{SICGAN results}
\label{RD_CycleGAN}
In the SICGAN, the objective is to balance improvements between the generator and discriminator networks. Ideally, the generator loss should decrease and stabilize as it consistently fools the discriminator, while the discriminator loss fluctuates around a midpoint, indicating high-quality, realistic image generation. The generator loss curves are calculated by averaging the total loss from~\eqref{eq:CycleGAN_TotLossSICGAN} over the number of samples $N$ \eqref{eq:SIGAN_averageLoss}, while the discriminator loss reflects the mean of the adversarial losses.
\begin{equation}
\resizebox{.8\hsize}{!}{$
\begin{aligned}
\mathcal{L}_{\mathrm{gen}\text{-}\mathrm{avg}}(G, F, D_{\mathcal{X}}, D_{\mathcal{Y}}) = \frac{1}{N} \sum_{i=1}^{N} \bigg(
& \mathcal{L}_{\mathrm{GAN}}(G, D_{\mathcal{Y}}, \mathcal{X}, \mathcal{Y}) + \mathcal{L}_{\mathrm{GAN}}(F, D_{\mathcal{X}}, \mathcal{Y}, \mathcal{X}) \\
&+ \lambda_{\mathrm{cyc}} \left( \mathcal{L}_{\mathrm{cyc}}^{G \rightarrow F} + \mathcal{L}_{\mathrm{cyc}}^{F \rightarrow G} \right) \\
&+ \lambda_{\mathrm{id}} \left( \mathcal{L}_{\mathrm{id}}^{G} + \mathcal{L}_{\mathrm{id}}^{F} \right) \bigg)
\end{aligned}$}
\label{eq:SIGAN_averageLoss}
\end{equation}

\begin{figure}[h!]
    \centering
    \includegraphics[width=0.65\columnwidth]{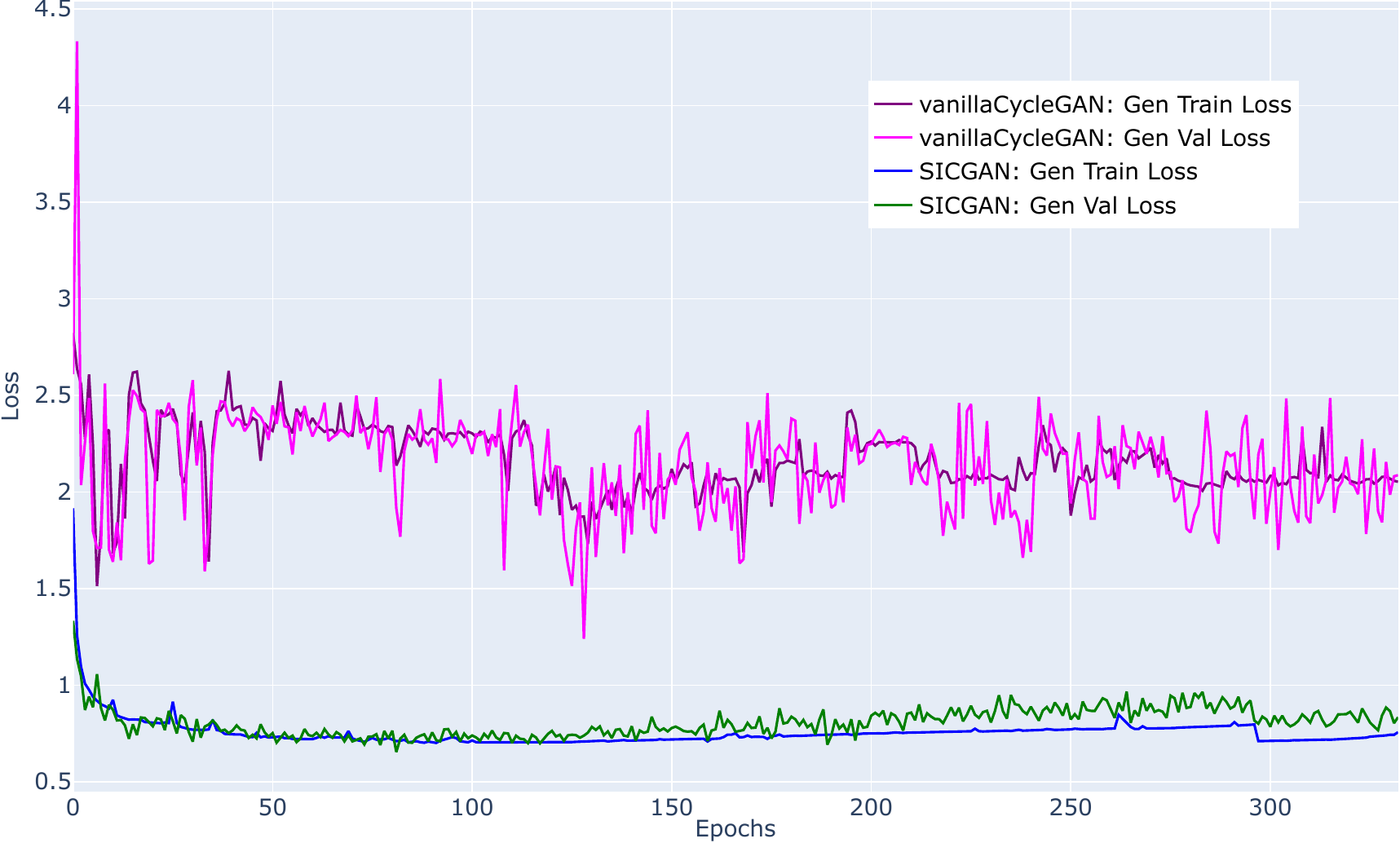}
    \caption{SICGAN and CycleGAN loss curves for the generators during training and validation in the IRB120 environment. Generator loss curves (purple and blue for training, pink and green for validation) combine both generators' losses.}
    \label{fig:sicganvscyclegan_losses}
\end{figure}

\begin{figure}[h!]
    \centering
    \subfloat[]{\includegraphics[width=0.2\columnwidth]{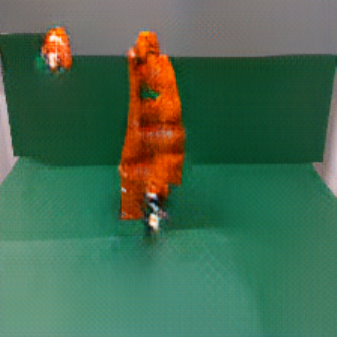}}
    \hspace{0.2\columnwidth}
    \subfloat[]{\includegraphics[width=0.2\columnwidth]{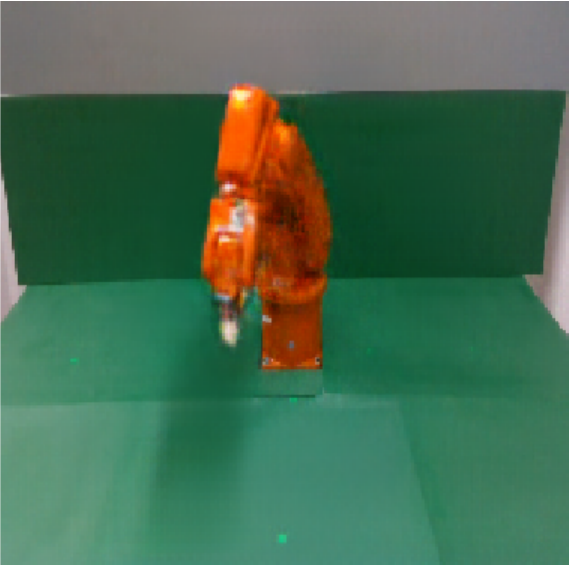}}
    \caption{Virtual-to-real observation translation with the CycleGAN (a) and the SICGAN (b) for the IRB120 environment.}
    \label{fig:sicganvscyclegan_img}
\end{figure}

The modifications introduced in the SICGAN to enhance the original CycleGAN architecture lead to significant improvements in both the training process and the quality of image-to-image translation between domains. As shown in Figure~\ref{fig:sicganvscyclegan_losses}, the loss curves obtained during the training of both architectures with the IRB120 dataset indicate faster convergence, greater stability, and an overall better performance of the SICGAN compared to the vanilla CycleGAN. Additionally, Figure~\ref{fig:sicganvscyclegan_img} illustrates the qualitative differences in translation results, where the best-performing SICGAN model generates real-synthetic images that more closely resemble the real domain, while the baseline CycleGAN exhibits noticeable visual artifacts and domain mismatches. These results confirm the effectiveness of the proposed enhancements in addressing the limitations of standard CycleGAN for domain adaptation in sim-to-real tasks.

\begin{figure}[h!]
    \centering
    \subfloat[]{\includegraphics[width=0.48\columnwidth]{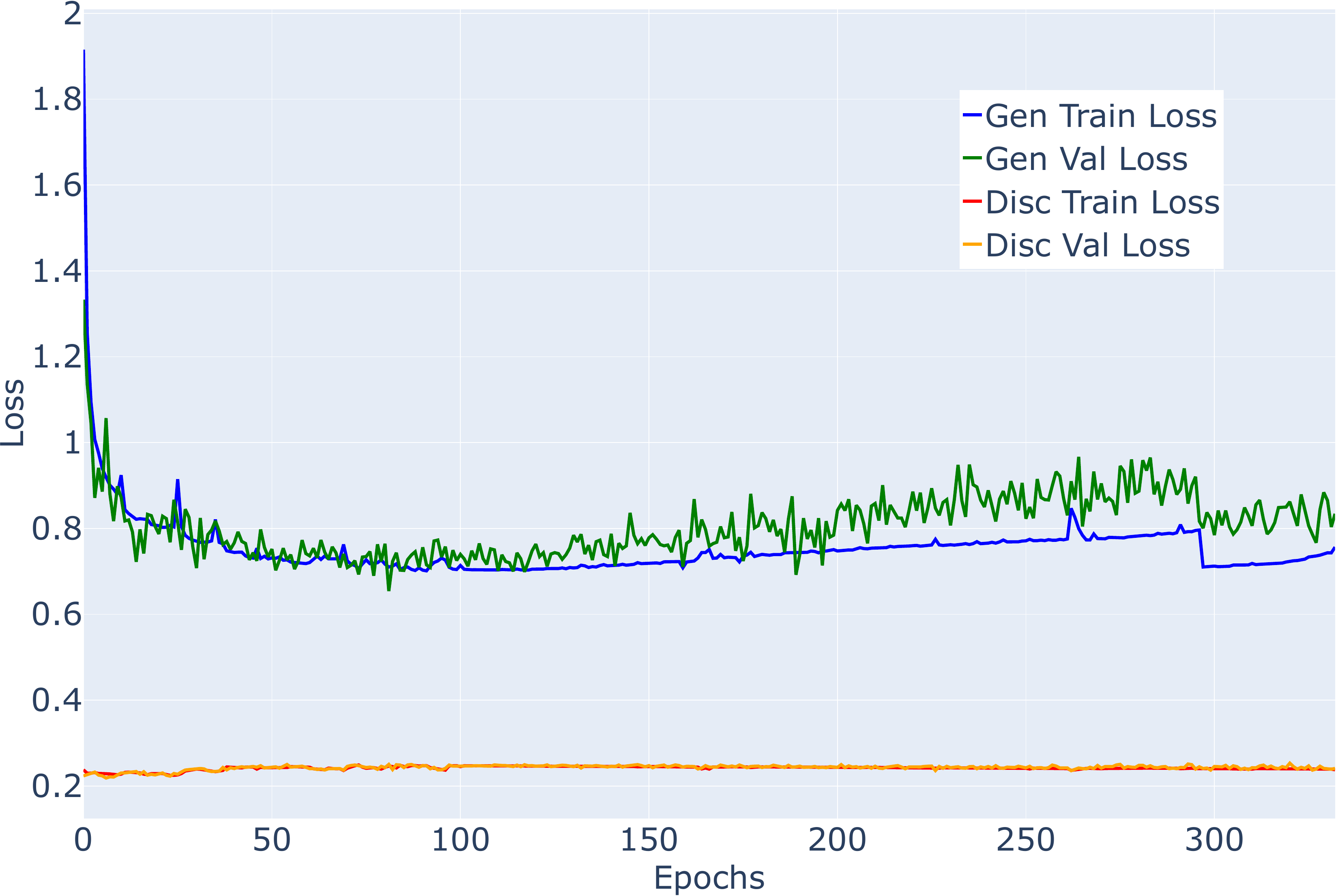}}
    \hspace{0.02\columnwidth}
    \subfloat[]{\includegraphics[width=0.48\columnwidth]{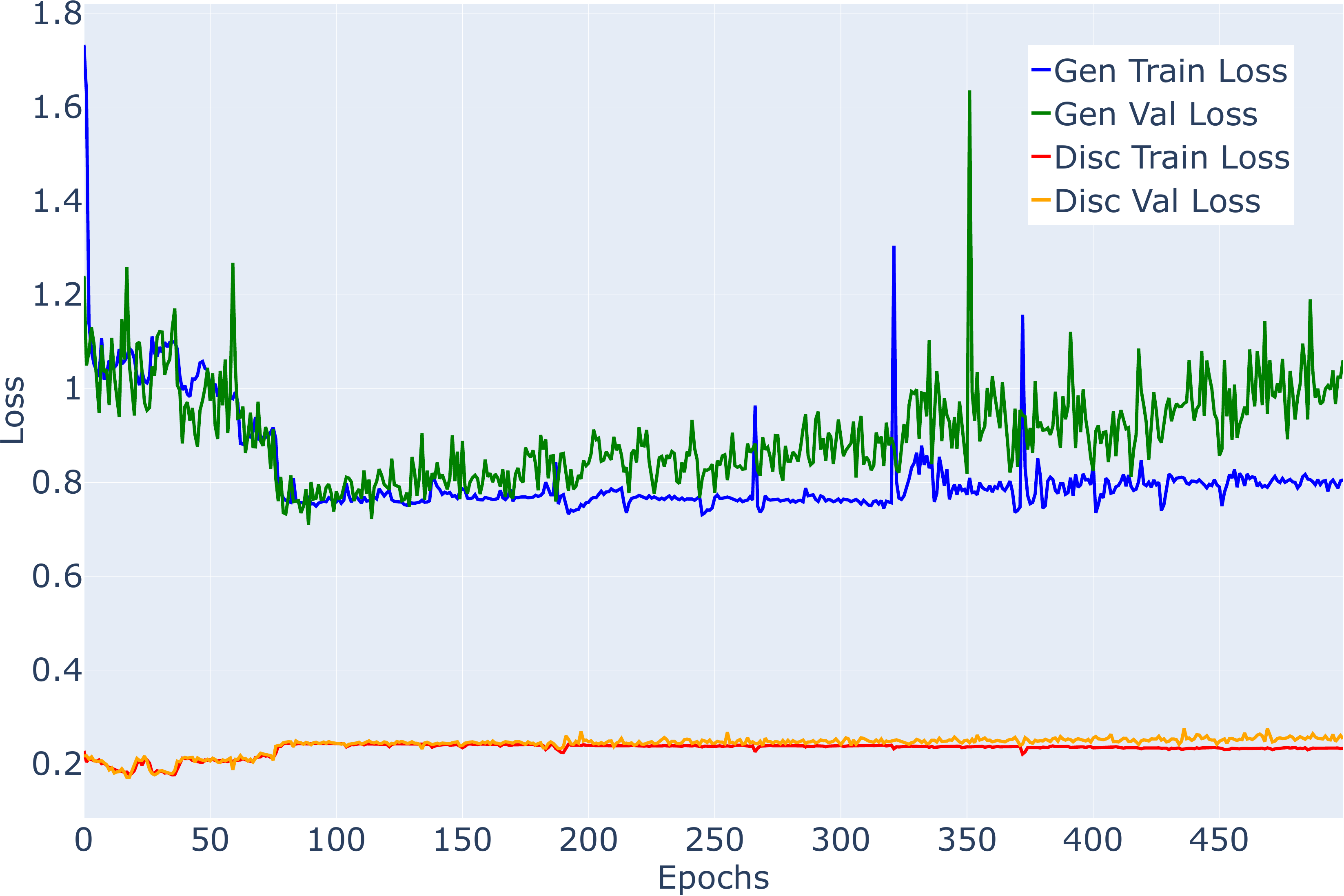}}
    \caption{SICGAN loss curves during training and validation in the IRB120 environment (a) and the UR3e environment (b). The IRB120 SICGAN is trained for approximately 300 epochs and 500 for the UR3e. Generator loss curves (blue for training, green for validation) combine both generators' losses, and similarly, discriminator loss curves (red for training, orange for validation) aggregate both discriminators.}
    \label{fig:sicganLosses}
\end{figure}

\begin{figure}[h!]
    \centering
    \subfloat[]{\includegraphics[width=0.15\columnwidth]{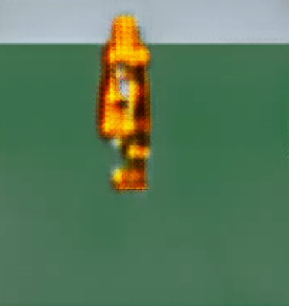}}
    \hspace{0.05\columnwidth}
    \subfloat[]{\includegraphics[width=0.159\columnwidth]{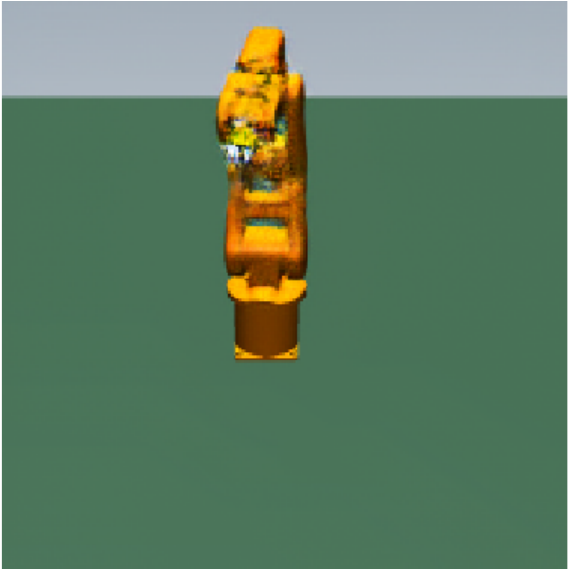}}
    \hspace{0.05\columnwidth}
    \subfloat[]{\includegraphics[width=0.157\columnwidth]{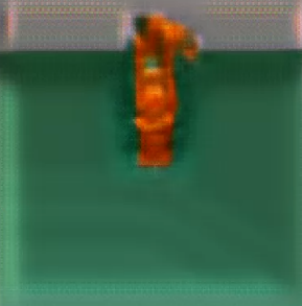}}
    \hspace{0.05\columnwidth}
    \subfloat[]{\includegraphics[width=0.162\columnwidth]{fig33.pdf}}
    \caption{Real-to-sim and sim-to-real observation translation at the initial SICGAN training stage (a)-(c) and in the final selected model (b)-(d) for the IRB120 environment.}
    \label{fig:obsTranslation_IRB120}
\end{figure}

After validating the SICGAN improvement with respect to the CycleGAN in the IRB120 setup, Figure~\ref{fig:sicganLosses} shows the training and testing losses for the SICGAN generator and discriminator in the IRB120 and UR3e environments. 
\FloatBarrier
In the IRB120 SICGAN training, while the generator's losses decrease rapidly, the discriminator's losses show minimal change. From epoch 150, generator losses begin to increase until a sharp drop around epoch 300 occurs, potentially indicating a partial mode collapse where the generator produces less diverse outputs, challenging the discriminator. This sudden drop suggests an improvement in generator performance, with the discriminator adjusting its response as losses fluctuate. Visual inspection of synthetic images from real-to-virtual and virtual-to-real translations confirms high-quality outputs (Figure~\ref{fig:obsTranslation_IRB120}), demonstrating the SICGAN's ability to learn effective representations despite loss instability. The selected model at epoch 114 has generator and discriminator training losses of 0.703 and 0.246, respectively, and validation losses of 0.719 and 0.246.

\begin{figure}[h!]
    \centering
    \subfloat[]{\includegraphics[width=0.15\columnwidth]{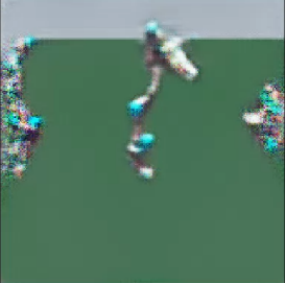}}
    \hspace{0.05\columnwidth}
    \subfloat[]{\includegraphics[width=0.15\columnwidth]{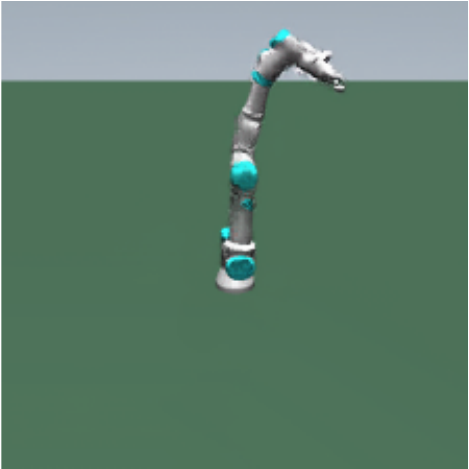}}
    \hspace{0.05\columnwidth}
    \subfloat[]{\includegraphics[width=0.15\columnwidth]{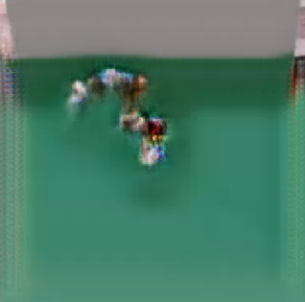}}
    \hspace{0.05\columnwidth}
    \subfloat[]{\includegraphics[width=0.15\columnwidth]{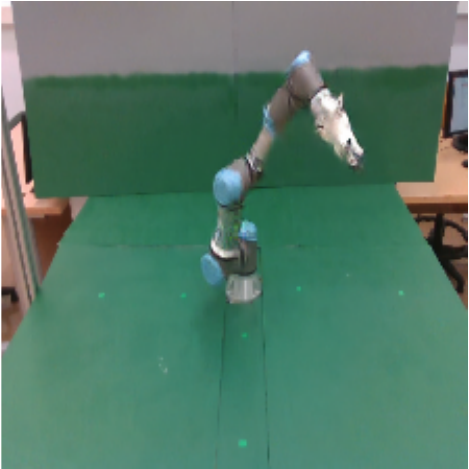}}
    \caption{Real-to-sim and sim-to-real observation translation at the initial SICGAN training stage (a)-(c) and in the final selected model (b)-(d) for the UR3e environment.}
    \label{fig:obsTranslation_UR3e}
\end{figure}

Analyzing the UR3e SICGAN training, the generator's losses decrease quickly, though more slowly than in the IRB120 case, with a slight decline in the discriminator’s losses, indicating the generator is improving but not yet fully deceiving the discriminator. Afterward, both losses stabilize, suggesting a balance where the generator produces reasonably realistic images and the discriminator maintains some differentiation ability. Past epoch 150, a gradual rise in the generator’s test loss points to challenges in generalizing to unseen data, while occasional loss spikes hint at possible training instabilities or mode collapse, similar to what was observed in the IRB120. Examining the generator's virtual-to-real translations, Figure~\ref{fig:obsTranslation_UR3e} demonstrates sufficient visual detail for our application. The selected model from epoch 249 achieved training losses of 0.741 (generator) and 0.237 (discriminator), with validation losses of 0.829 and 0.251, respectively.

\begin{figure}[h!]
    \centering
    \includegraphics[width=0.65\columnwidth]{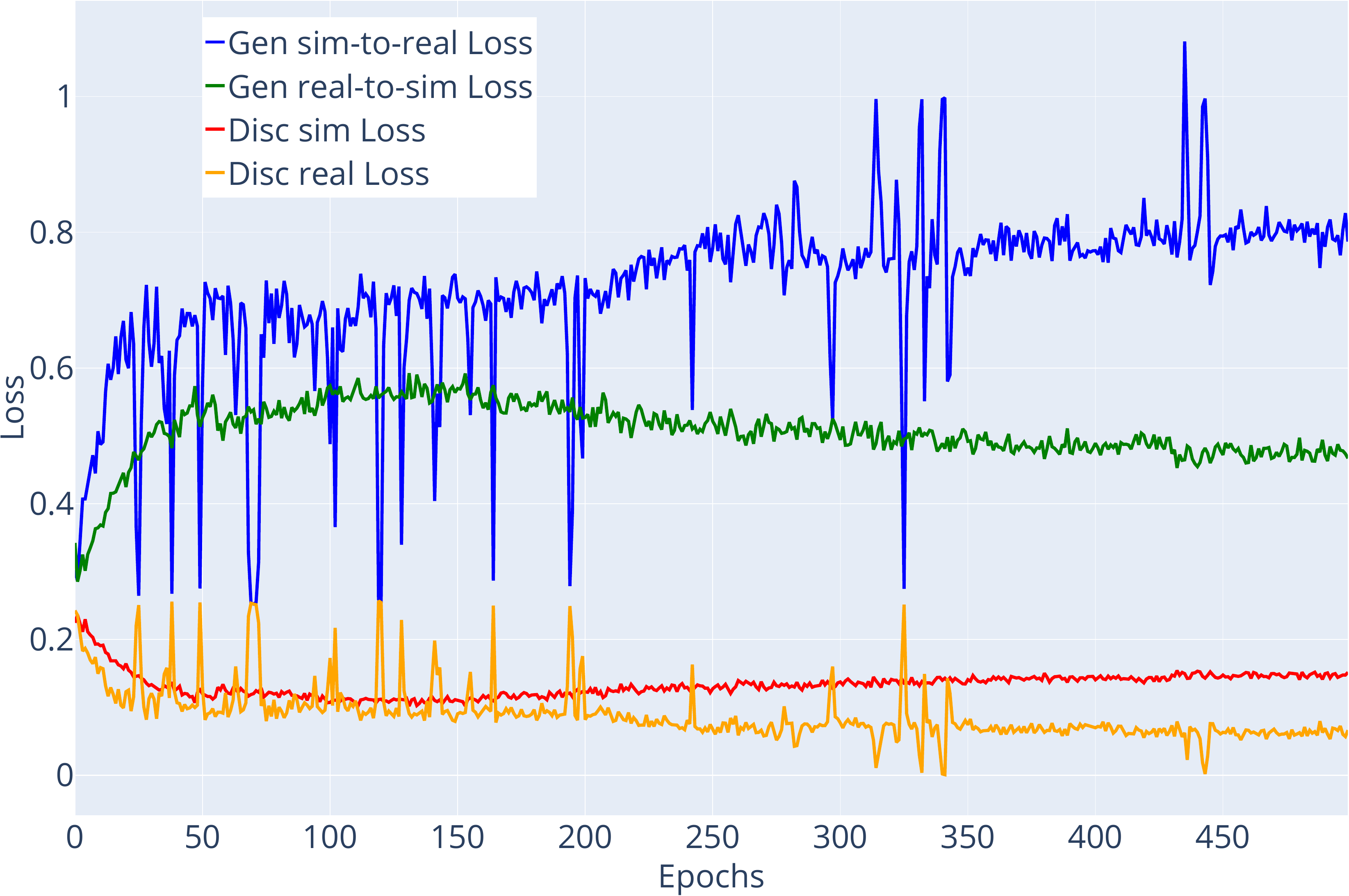}
    \caption{UVCGANv2 loss curves for the generators and discriminators in the IRB120 environment.}
    \label{fig:uvcgan_losses}
\end{figure}

\begin{figure}[h!]
    \centering
    \centering
    \subfloat[]{\includegraphics[width=0.2\columnwidth]{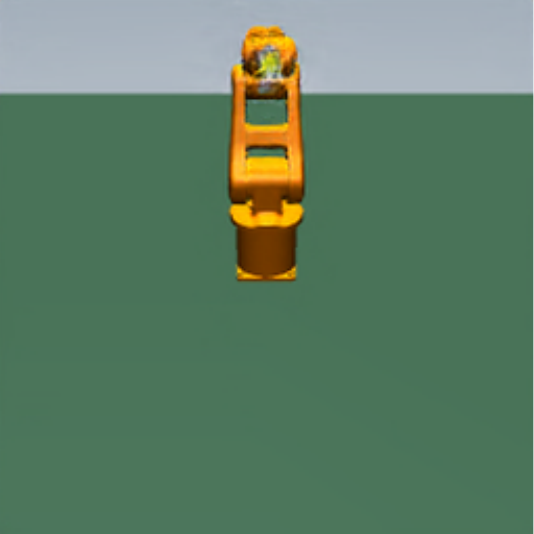}}
    \hspace{0.2\columnwidth}
    \subfloat[]{\includegraphics[width=0.2\columnwidth]{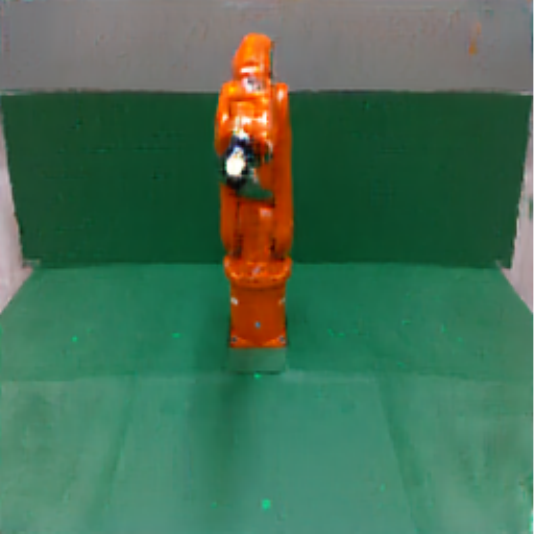}}
    \caption{Real-to-sim (a) and sim-to-real (b) observation translation using the UVCGANv2 best model for the IRB120 environment.}
    \label{fig:obsTranslation_IRB120_uvcganv2}
\end{figure}

Regarding the comparison of our SICGAN with other state-of-the-art methods, this work includes an evaluation against UVCGANv2, introduced in Section~\ref{RW}. Its training procedure involves two stages: a pre-training phase of the generators using a self-supervised inpainting task, followed by the full training of the complete architecture. In our study, we reproduced the training procedure described in the original UVCGANv2 paper using our own IRB120 dataset, following the same configuration and hyperparameters used for the AFHQ dataset~\citep{Torbunov2023_b}. Figure~\ref{fig:uvcgan_losses} presents the loss curves of the model throughout the training process, illustrating its convergence behavior under our experimental conditions. The generator losses for both, the sim-to-real and real-to-sim directions, exhibit a progressive increase over time, which may indicate instability or divergence in the generator training. This behavior suggests that the generators struggle to maintain consistent performance as training progresses, potentially due to a mismatch in convergence rates between generators and discriminators or an overfitting effect in the discriminators. In contrast, the discriminator losses remain relatively low and stable throughout the training process. Figure~\ref{fig:obsTranslation_IRB120_uvcganv2} shows the real-to-sim and sim-to-real observations.

\begin{figure}[h!]
    \centering
    \subfloat[]{\includegraphics[width=0.45\columnwidth]{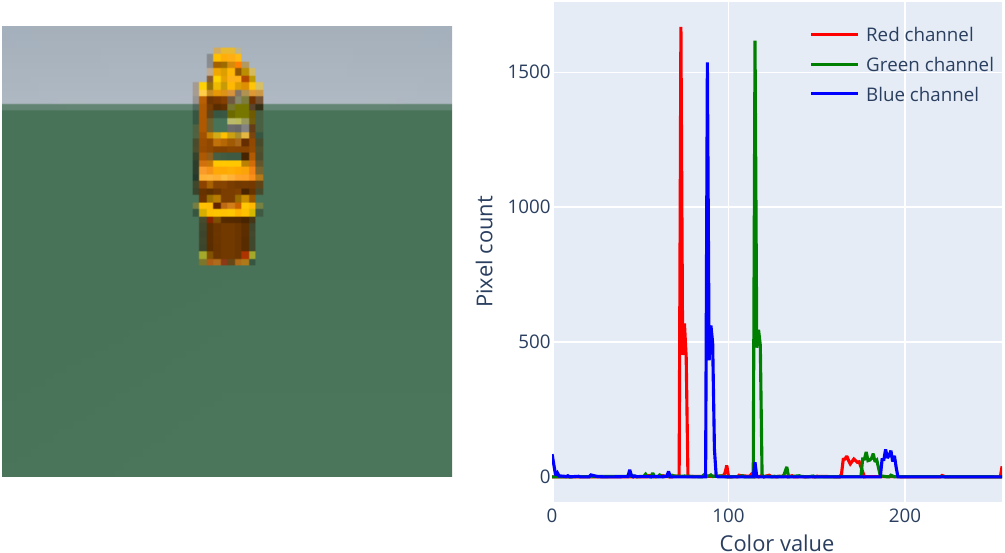}}
    \hspace{0.05\columnwidth}
    \subfloat[]{\includegraphics[width=0.45\columnwidth]{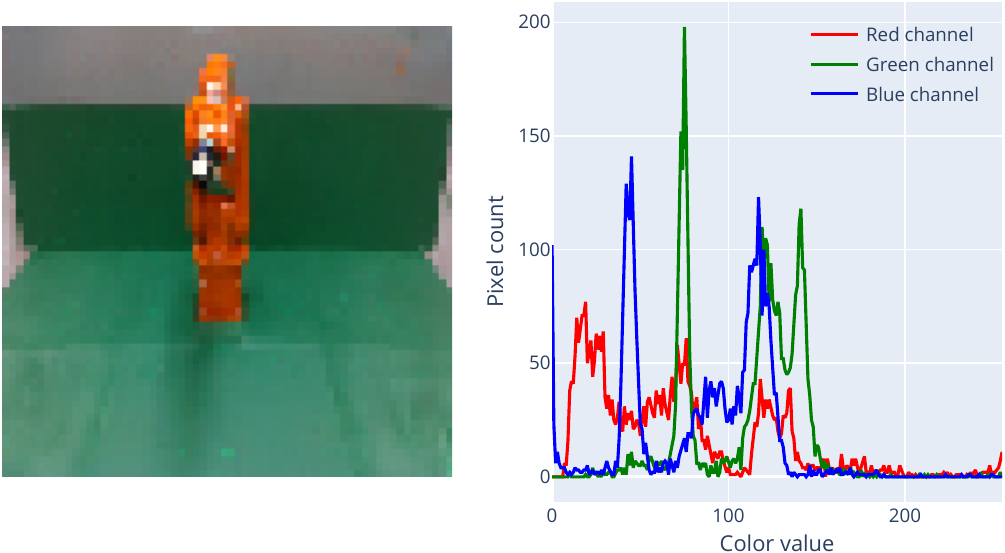}}
    \hspace{0.05\columnwidth}
    \subfloat[]{\includegraphics[width=0.45\columnwidth]{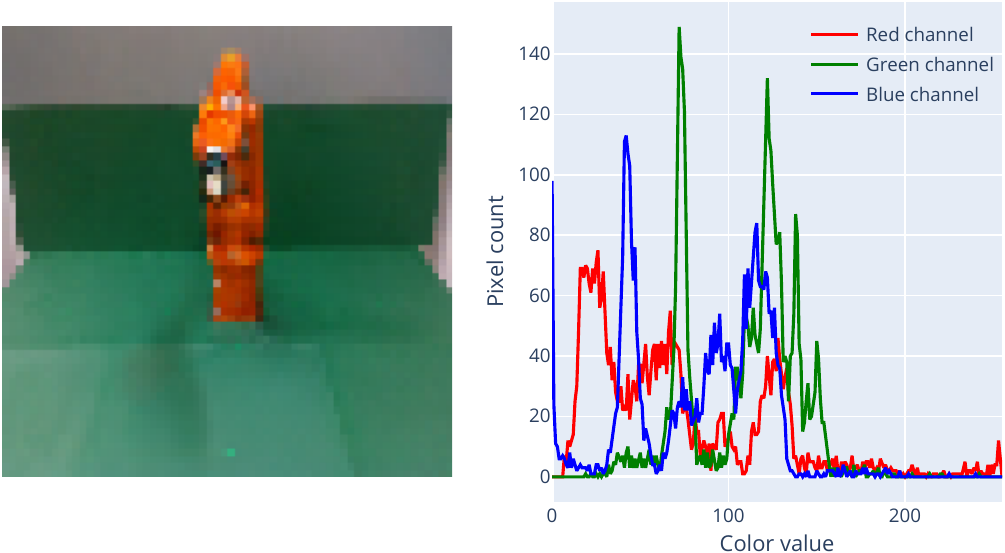}}
    \hspace{0.05\columnwidth}
    \subfloat[]{\includegraphics[width=0.45\columnwidth]{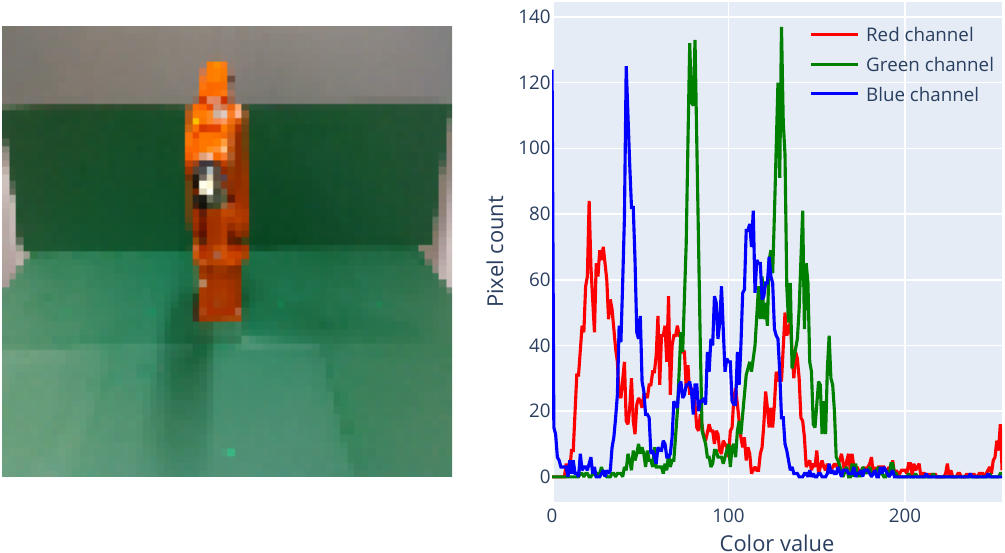}}
    \caption{RGB histograms for raw virtual (a), UVCGANv2 real-synthetic (b), SICGAN real-synthetic (c), and real (d) observations.}
    \label{fig:ObsHistograms}
\end{figure}

Figure~\ref{fig:ObsHistograms} compares the RGB histograms for (a) raw virtual, (b) UVCGANv2 real-synthetic, (c) SICGAN real-synthetic, and (d) real observations in the IRB120 setup. As shown in the histogram distributions, both UVCGANv2 and SICGAN are capable of generating images from virtual observations that closely resemble those from the real domain. To provide a more objective comparison of the visual similarity and how well each model aligns with the real distribution at pixel level, Table~\ref{tab:RGBSimilarityMetrics} reports the Wasserstein Distance computed over the RGB channel distributions.
\begin{table}[h!]
\centering
\caption{Wasserstein Distance between the real image, used as the reference, and the virtual and translated observations.}
\label{tab:RGBSimilarityMetrics}
\renewcommand*{\arraystretch}{1.2}
\resizebox{0.6\columnwidth}{!}{%
\begin{tabular}{>{\centering\arraybackslash}m{4cm} 
                >{\centering\arraybackslash}m{2cm} >{\centering\arraybackslash}m{2cm} >{\centering\arraybackslash}m{2cm}}
    \toprule
    \textbf{Observation} & \textbf{R} & \textbf{G} & \textbf{B} \\
    \midrule[2pt]
    Virtual & 0.0055 & 0.0050 & 0.0050 \\
    UVCGANv2 & 0.0003 & 0.0007 & 0.0009 \\
    SICGAN & \textbf{0.0002} & \textbf{0.0003} & \textbf{0.0002} \\
    \bottomrule
\end{tabular}%
}
\end{table}

The quantitative results presented in Table~\ref{tab:RGBSimilarityMetrics} highlight the strong performance of both UVCGANv2 and SICGAN in translating virtual observations into realistic representations. While both models significantly outperform the baseline virtual observations, SICGAN consistently achieves lower WD values. These results indicate a slightly closer alignment between SICGAN-generated images and the real domain distribution, which is critical for effective sim-to-real transfer in robotics. Moreover, SICGAN achieves this performance with a more lightweight architecture and simpler training procedure compared to the more complex, two-stage training process required by UVCGANv2. Based on this favorable trade-off between performance, architectural simplicity, and training efficiency, we choose to focus on SICGAN for the continued development of our methodology, this is, generating real-synthetic observations for training the agent in the virtual environment, and to enable zero-shot deployment in the real robot workspace.

\subsection{DRL agent results in the virtual environment}
\label{RD_DRLagent}
\begin{figure}[h!]
    \centering
    \includegraphics[width=0.65\columnwidth]{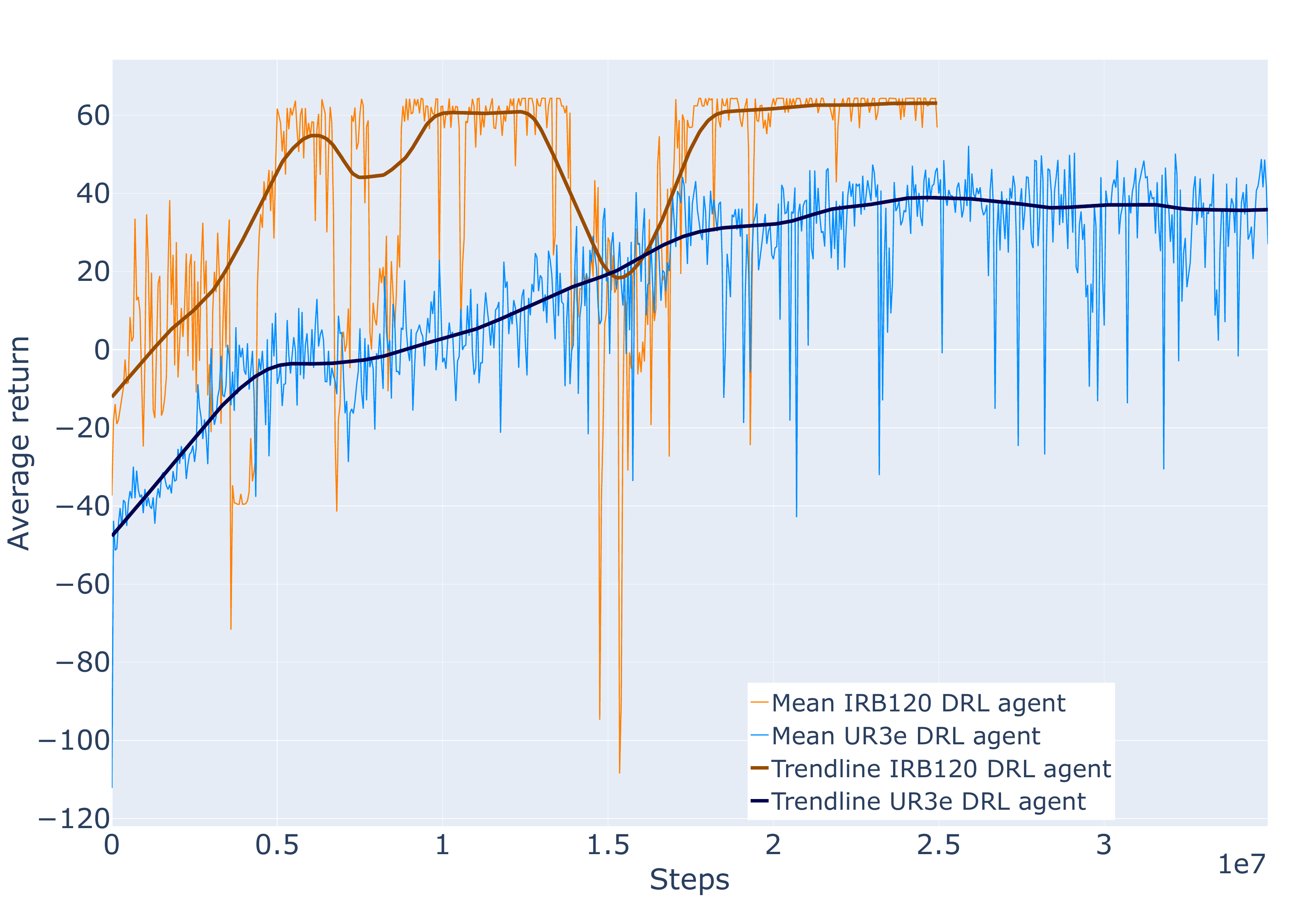}
    \caption{Average returns from interim evaluations of agents trained on real-synthetic observations in the IRB120 environment (orange) and UR3e environment (blue) over 35M training steps (note the scaling factor of 1e7 at the bottom-right corner). The training process of the IRB120 agent is stopped earlier due to the good results achieved.}
    \label{fig:DRLmodelsTrained}
\end{figure}

Figure~\ref{fig:DRLmodelsTrained} presents the training curves for the IRB120 (orange) and UR3e (blue) agents. \revd{The 35M training steps correspond to 161 training hours using only 8 workers due to the GPU RAM limitations.} Both agents reach a steady state, but the IRB120 agent learns faster and achieves a higher final average return. The fact that the agent trained in the UR3e setup needs more experience to learn and reaches a lower performance might be explained due to the differences in the arm morphology, as the movements of the UR3e are more challenging to understand because of the offsets between joints. The best model for the IRB120 environment is obtained at step 6.35M, achieving \qty{100}{\percent} accuracy in the post-training virtual evaluation, while for the UR3e the best model is found at step 25M with a \qty{90}{\percent} accuracy.

\subsection{Zero-shot transfer}
\label{RD_ZeroShot}
As shown in Figure~\ref{fig:ObsHistograms}, the raw virtual observation presents significant visual mismatches leading to a zero-shot transfer accuracy of only \qty{15.8}{\percent}, which is obtained using solely the two CNN layers of the agent's architecture. This baseline serves to evaluate whether our proposed methodology effectively improves the agent's performance in the real scenario. Conversely, the SICGAN translated observations closely match the real observation distribution, suggesting good zero-shot results.

\begin{figure}[h!]
    \centering
    \includegraphics[width=0.65\columnwidth]{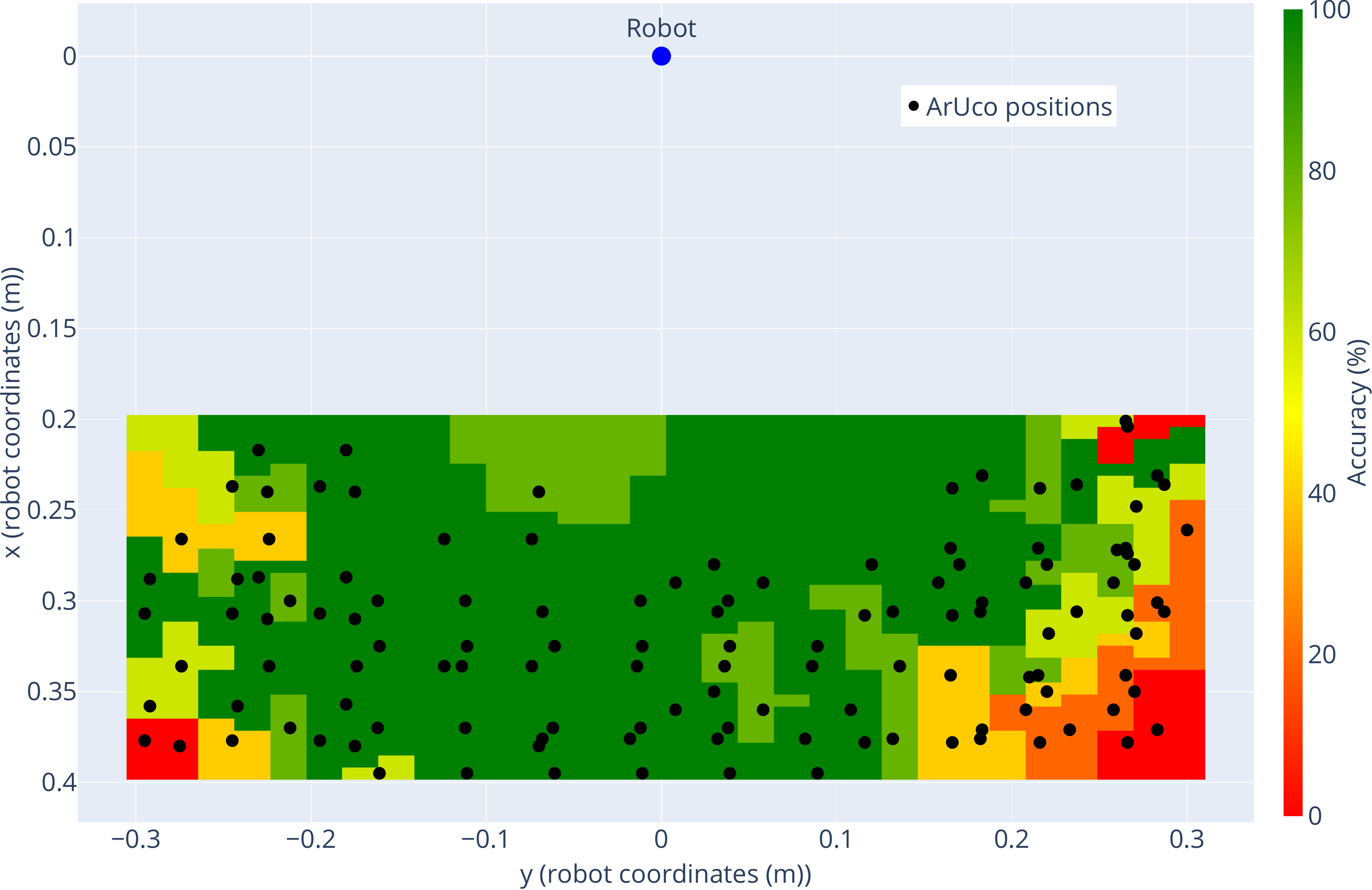}
    \caption{Accuracy heatmap for the DRL agent trained with real-synthetic observations, evaluated with AR targets in the IRB120 real environment. The $x$ and $y$ axes represent the robot's coordinates, and the color gradient indicates accuracy percentages.}
    \label{fig:heatmap_arucos_IRB120}
\end{figure}

\begin{figure}[h!]
    \centering
    \includegraphics[width=0.9\columnwidth]{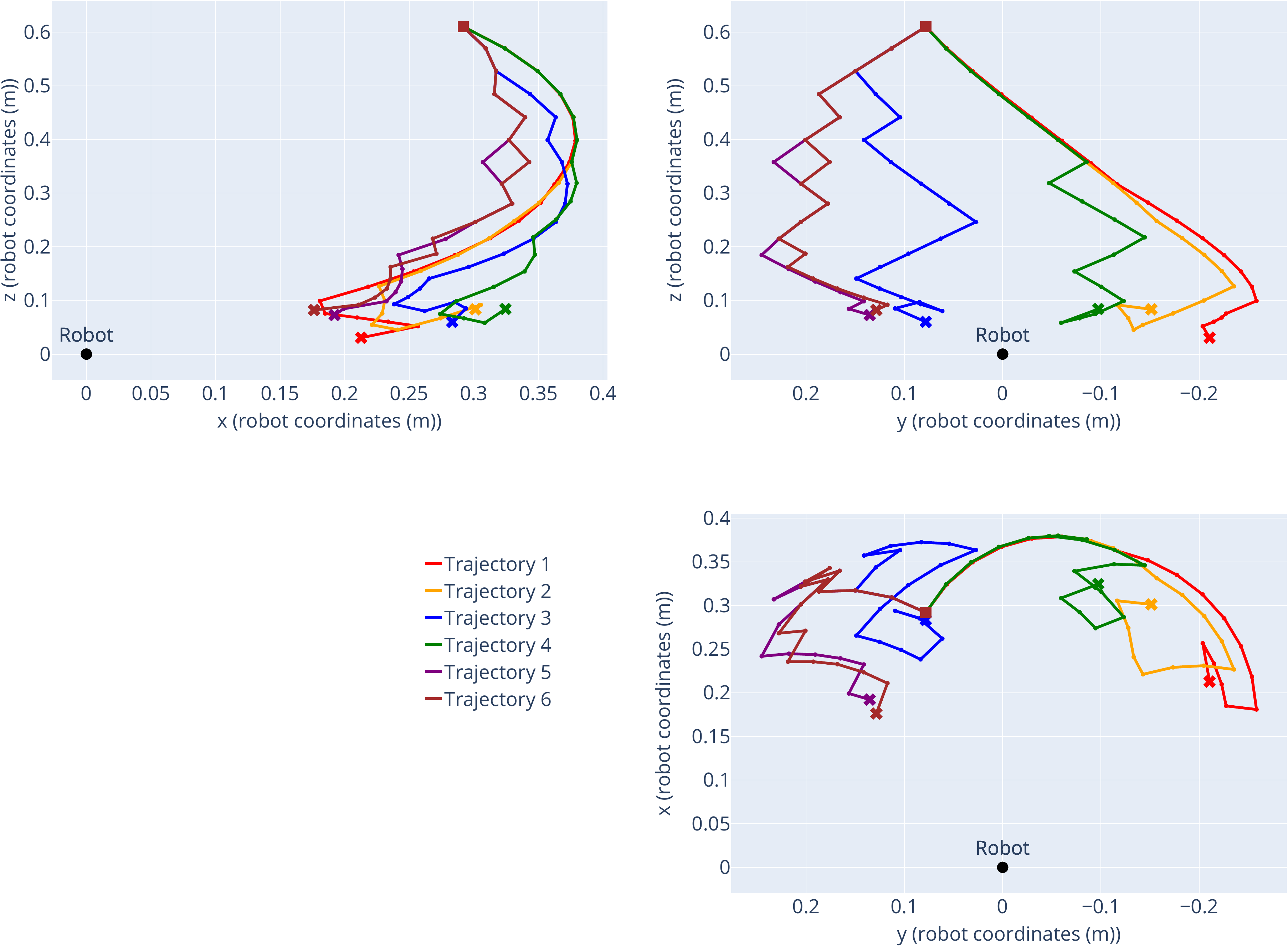}
    \caption{3D robot trajectories in the IRB120 real environment projected onto plan, front, and side views, obtained through zero-shot transfer. The red and orange trajectories correspond to targets in the negative $y$ region, the blue and green to central targets, and the purple and brown to targets in the positive $y$ region. All trajectories successfully reach the target except the fifth. Square markers indicate initial positions, while crosses denote final locations.}
    \label{fig:trajectories}
\end{figure}

Figure~\ref{fig:heatmap_arucos_IRB120} presents the accuracy results distributed across the workspace. The agent achieves a success rate of \qty{100}{\percent} for most ArUco tag positions. Although some areas (e.g., positive $y$ coordinates and the far right) have lower accuracies, these occur primarily in regions that are in the workspace limits. As illustrated, the workspace is considerably large, extending towards the boundary regions of the robot's reach. In these outer areas, the manipulator can adopt configurations for which the kinematic solution becomes unfeasible. Consequently, it is more challenging for the agent to accurately reach target positions located near these limits. Although this boundary region is not of particular interest under typical operating conditions ---where position density is naturally higher in the central area of the workspace--- we deliberately included it in our experiments to assess the limits of the agent's performance. Analyzing the robot gripper trajectories for six episodes, Figure~\ref{fig:trajectories} shows that the SICGAN-trained agent adapts well to target locations.

\begin{figure}[h!]
    \centering
    \includegraphics[width=0.65\columnwidth]{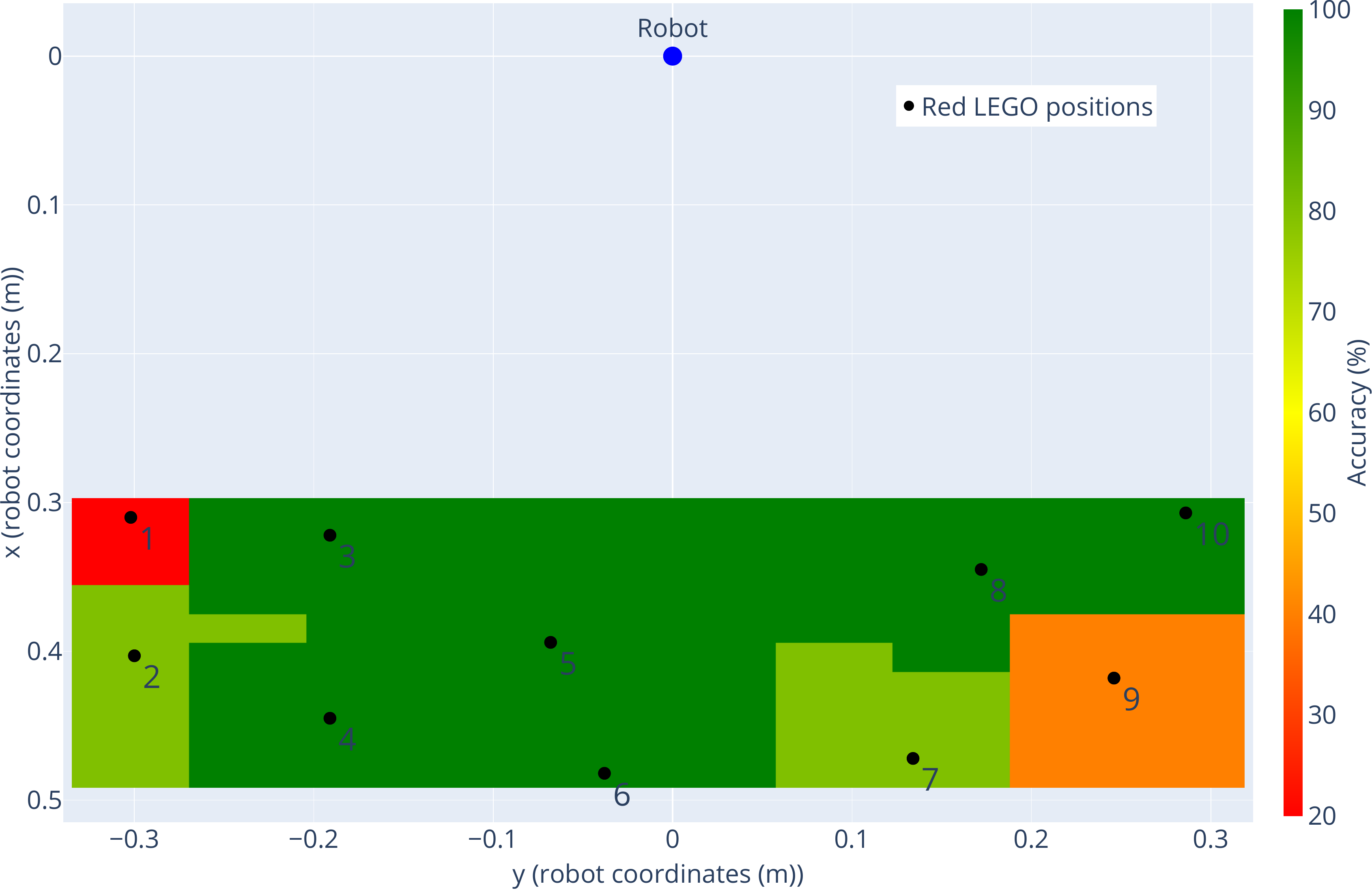}
    \caption{Accuracy heatmap for the DRL agent trained on real-synthetic observations, evaluated with the red LEGO\textsuperscript{\textregistered} cube in the IRB120 real environment. The $x$ and $y$ axes represent the robot's coordinates, and the color gradient indicates the accuracy percentage.}
    \label{fig:heatmap_redCube_IRB120}
\end{figure}

Evaluating the agent performance with the real targets, the zero-shot results with the red LEGO\textsuperscript{\textregistered} cube (Figure~\ref{fig:heatmap_redCube_IRB120}) reflect a success distribution similar to that with ArUco tags, validating the use of AR targets for efficient performance evaluation. The agent reached the target at least once in all ten tested positions, confirming its generalization capability towards a real target similar to the ones used in training.
\begin{table}[t]
    \centering
    \renewcommand*{\arraystretch}{1.0}
    \scriptsize
    \caption{Zero-shot accuracy for the yellow and blue LEGO\textsuperscript{\textregistered} cubes, and the red and white mug at different positions in the IRB120 environment.}
    \label{tab:accuracy_realTargets_IRB120}
    \resizebox{0.8\columnwidth}{!}{ 
    \begin{tabular}{c>{\centering\arraybackslash}m{3cm}>{\centering\arraybackslash}m{2cm}>{\centering\arraybackslash}m{1.5cm}>{\centering\arraybackslash}m{3cm}}
        \toprule
        \textbf{Position ID} & \textbf{\boldmath$(x, y)$ coordinates \linebreak (m)} & \textbf{Yellow cube} & \textbf{Blue cube} & \textbf{Red and white mug} \\ 
        \midrule
        3 & (0.322, --0.191) & \textbf{100\%} & 0\% & \textbf{100\%} \\ 
        4 & (0.445, --0.191) & \textbf{100\%} & 0\% & 40\% \\ 
        5 & (0.394, --0.068) & \textbf{100\%} & 20\% & \textbf{100\%} \\ 
        8 & (0.345, 0.172) & \textbf{100\%} & \textbf{100\%} & \textbf{100\%} \\ 
        \bottomrule
    \end{tabular}
    }
\end{table}

The agent's generalization capability is further tested with the yellow and blue LEGO\textsuperscript{\textregistered} cubes, and the red and white mug. The results gathered in Table~\ref{tab:accuracy_realTargets_IRB120} focus on the four positions where this agent achieved \qty{100}{\percent} success with the red cube. The agent maintains perfect performance with the yellow cube and decreases only in one position for the red and white mug. The worst performance is observed for the blue cube, which also happens when we evaluate the agent in the virtual environment. We suggest that this issue may arise from the learned CNN filters, as the target's RGB code during training (1.0, 0.0, 0.0) does not activate the blue channel. To address this, future work could explore incorporating DR for the target color during the virtual training phase, enhancing the variability seen by the agent.

\begin{figure}[h!]
    \centering
    \includegraphics[width=0.65\columnwidth]{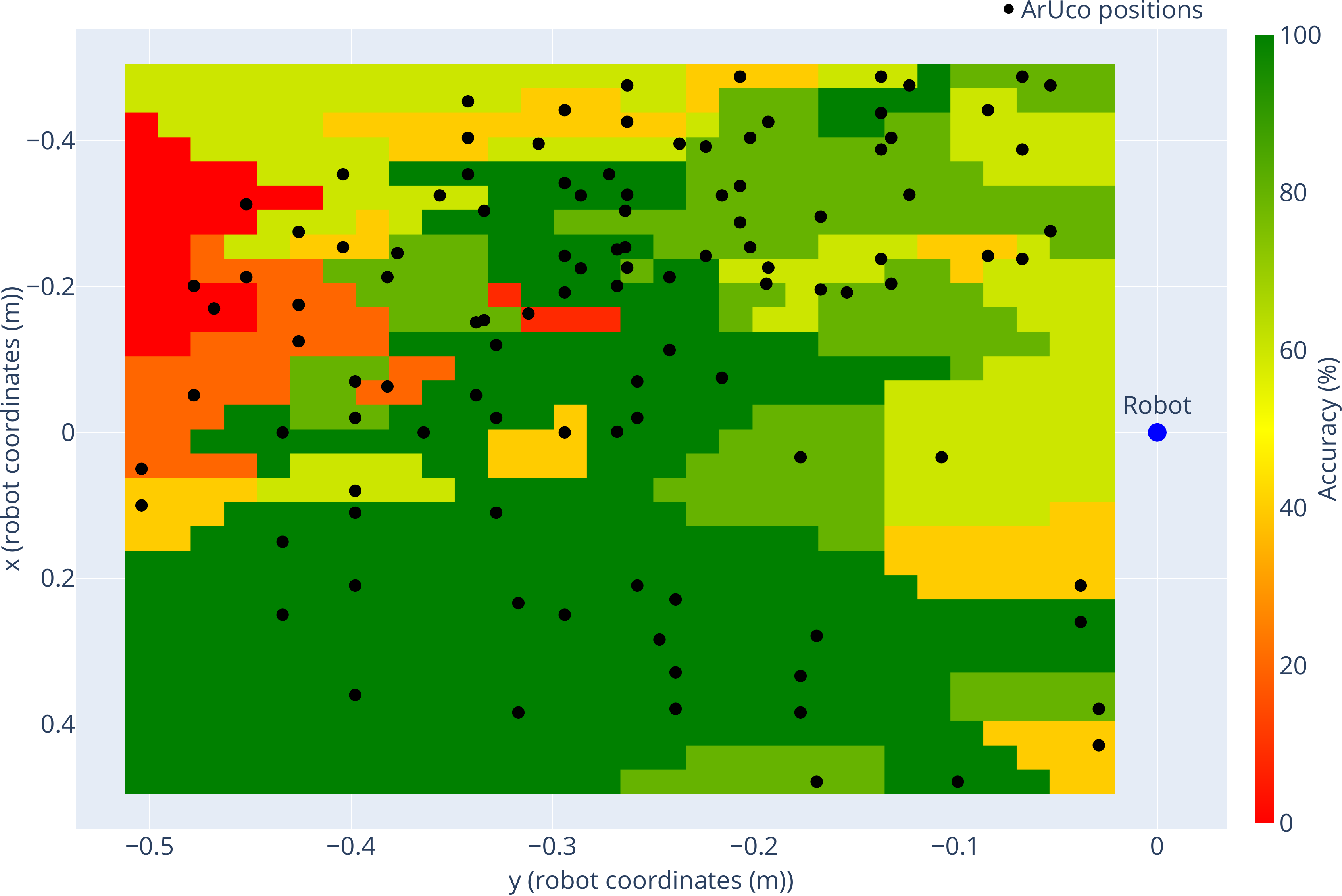}
    \caption{Accuracy heatmap for the DRL agent trained on real-synthetic observations, evaluated with AR targets in the UR3e real environment. The $x$ and $y$ axes represent the robot's coordinates, and the color gradient indicates accuracy percentages.}
    \label{fig:heatmap_arucos_UR3e}
\end{figure}

To finish validating the designed pipeline, Figure~\ref{fig:heatmap_arucos_UR3e} shows the heatmap with the accuracy percentages achieved in the zero-shot transfer of the virtually trained agent over the UR3e environment using AR targets. Higher accuracies are observed near positive $x$ coordinates and central positions, with a slight decline as targets move toward the workspace boundaries. This drop may result from increased complexity due to larger displacements or target orientation changes, and is consistent with findings from the zero-shot transfer results in the IRB120 environment.

In this case, the experiments with real targets are conducted in five previously unseen positions. Table~\ref{tab:accuracy_realTargets_UR3e} summarizes the mean accuracies for each position and target. The results demonstrate the DRL agent's ability to generalize effectively. As with the IRB120, the lowest success rates were observed with the blue cube likely due to the same problem with the state representation.

Overall, this results validate the proposed pipeline for addressing the sim-to-real problem, demonstrating robust performance in a wide range of scenarios. Additionally, the outcomes suggest that the robot's geometry and fixed camera position may affect accuracy in specific regions, which could inform future virtual training setups to better cover challenging areas.
\begin{table}[t]
    \centering
    \renewcommand*{\arraystretch}{1.0}
    \scriptsize
    \caption{Zero-shot accuracy for the yellow and blue LEGO\textsuperscript{\textregistered} cubes, and the red and white mug at different positions in the UR3e environment.}
    \label{tab:accuracy_realTargets_UR3e}
    \resizebox{0.8\columnwidth}{!}{ 
    \begin{tabular}{>{\centering\arraybackslash}m{3cm}>{\centering\arraybackslash}m{1.5cm}>{\centering\arraybackslash}m{2cm}>{\centering\arraybackslash}m{1.5cm}>{\centering\arraybackslash}m{3cm}}
        \toprule
        \textbf{\boldmath$(x, y)$ coordinates \linebreak (m)} & \textbf{Red cube} & \textbf{Yellow cube} & \textbf{Blue cube} & \textbf{Red and white mug} \\ 
        \midrule
        (--0.385, --0.228) & \textbf{100\%} & \textbf{100\%} & 60\% & \textbf{100\%} \\ 
        (--0.285, --0.378) & \textbf{100\%} & \textbf{100\%}  & 80\% & \textbf{100\%} \\ 
        (0.011, --0.437) & \textbf{100\%} & \textbf{100\%} & 60\% & 80\%\\ 
        (0.319, --0.298) & \textbf{100\%} & \textbf{100\%} & 40\% & \textbf{100\%}\\ 
        (0.403, --0.102) & \textbf{100\%} & \textbf{100\%} & 20\% & 80\%\\
        \bottomrule
    \end{tabular}
    }
\end{table}

\FloatBarrier
\section{Conclusion}
\label{Conclusion}
We propose a zero-shot transfer methodology based on SICGAN, an original CycleGAN architecture that leverages DA for image-to-image translation. After training and validating a SICGAN model, virtual observations are translated into real-synthetic images to train the agents in simulation. These agents, which achieve success rates between 90 and \qty{100}{\percent}, are then deployed directly in the real environment using raw camera inputs. This approach avoids hardware limitations during training and opens the door to future work exploring the inverse translation (real-to-virtual) for comparison. To improve the SICGAN's training efficiency, a systematic hyperparameter optimization is advisable. Additionally, since our SICGAN model has been trained on a fixed background, its generalization may be limited when background conditions change. Addressing this limitation would involve training with neutral backgrounds and background subtraction during inference, or applying DR to enhance robustness.

The evaluated agents achieved excellent zero-shot performance across most of the workspaces, showing reliable trajectory planning and consistent target localization with accuracies above \qty{95}{\percent}, except for a few edge cases. Performance slightly degraded in scenarios involving the blue LEGO\textsuperscript{\textregistered} cube, likely due to incomplete state representation. Future work will incorporate DR during training or explore multi-agent training across tasks in simulation prior to deployment. Furthermore, regarding the application domain, once the methodology has been validated on two different robotic manipulators, future research can focus on extending the application to other industrial settings and enhance the agent's generalization capability by including DR during the learning process.

Overall, the results validate the proposed zero-shot pipeline as an effective sim-to-real solution for industrial scenarios. While it requires training an additional SICGAN model, it removes the need for real-world training and the hyperparameter uncertainty common in few-shot approaches.

\section*{Author contributions}
\noindent \textbf{Luc\'{i}a G\"{u}itta-L\'{o}pez}: Conceptualization and design of this study, Methodology, Formal analysis and investigation, Software, Data curation, Writing - original draft preparation, Writing - review and editing. \linebreak \textbf{Lionel G\"{u}itta-L\'{o}pez}: Conceptualization and design of this study, Methodology, Formal analysis and investigation, Software. \linebreak \textbf{Jaime Boal}: Conceptualization and design of this study, Methodology, Formal analysis and investigation, Supervision, Writing - review and editing.\linebreak \textbf{\'{A}lvaro J. L\'{o}pez-L\'{o}pez}: Conceptualization and design of this study, Methodology, Formal analysis and investigation, Supervision, Writing - review and editing.

\section*{Declarations}
\begin{itemize}
	\item Funding: This research did not receive any specific grant from funding agencies in the public, commercial, or not-for-profit sectors.
	\item Conflict of interest/Competing interests: All authors certify that they have no affiliations with or involvement in any organization or entity with any financial interest or non-financial interest in the subject matter or materials discussed in this manuscript.
	\item Availability of data and materials: Researchers or interested parties are welcome to contact the corresponding author L.G-L. for further explanation.
	\item Code availability: Researchers or interested parties are welcome to contact the corresponding author L.G-L. for further explanation.
\end{itemize} 

\bibliographystyle{elsarticle-harv} 
\bibliography{Manuscript_bibliography_r1}

\end{document}